%% file: acl.tex
\pdfoutput=1

\PassOptionsToPackage{table,dvipsnames}{xcolor}

\documentclass[11pt]{article}

\usepackage{acl}
\usepackage{times}
\usepackage{latexsym}

\usepackage[T1]{fontenc}
\usepackage[utf8]{inputenc}
\usepackage{microtype}
\usepackage{inconsolata}
\usepackage{textcomp}
\usepackage{CJK} 

\usepackage{amsmath}
\usepackage{amssymb}
\usepackage{amsfonts}
\usepackage{bm}
\usepackage{wasysym}
\usepackage{eufrak}
\usepackage{pifont}

\usepackage{algorithm}
\usepackage{algpseudocode}

\usepackage{booktabs}
\usepackage{multirow}
\usepackage{makecell}
\usepackage{tabularx}
\usepackage{arydshln}
\usepackage{balance}
\usepackage{enumitem}
\usepackage{graphicx}

\usepackage{xcolor} 
\usepackage{colortbl}
\usepackage{tcolorbox}
\tcbuselibrary{skins}

\definecolor{graybg}{rgb}{0.9,0.9,0.9}
\newcommand{\gain}[1]{\textcolor[rgb]{0.0, 0.5, 0.0}{$_{\uparrow\textbf{#1}}$}}
\newcommand{\loss}[1]{\textcolor[rgb]{0.8, 0.0, 0.0}{$_{\downarrow\textbf{#1}}$}}

\usepackage{url}

\definecolor{MyDarkBlue}{rgb}{0,0.08,0.45} 
\definecolor{MyLightBlue}{RGB}{235,245,255}

\newcommand{\cmark}{\ding{51}} 
\newcommand{\xmark}{\ding{55}} 

\newcommand{\zqh}{\color{black}}

\algnewcommand{\LineComment}[1]{\Statex \(\triangleright\) \textit{#1}}
\algnewcommand{\RightComment}[1]{\Comment{\textit{#1}}}

\title{\textbf{Try, Check and Retry}: A Divide-and-Conquer Framework for Boosting Long-context Tool-Calling Performance of LLMs}

\author{%
  Kunfeng~Chen$^{1}$,
  Qihuang~Zhong$^{1}$,
  \textbf{Juhua~Liu}$^{1}$,
  \textbf{Bo~Du}$^{1}$,
  \textbf{Dacheng~Tao}$^{2}$ \\
  \fontsize{9.0pt}{\baselineskip}\selectfont $^{1}$ School of Computer Science, National Engineering Research Center for Multimedia Software, Institute of Artificial Intelligence\\ 
  \fontsize{9.0pt}{\baselineskip}\selectfont  and Hubei Key Laboratory of Multimedia and Network Communication Engineering, Wuhan University, China \\
  \fontsize{9.0pt}{\baselineskip}\selectfont $^{2}$ Nanyang Technological University, Singapore \\
   \fontsize{9.0pt}{\baselineskip}\selectfont \texttt{\{chenkunfeng, zhongqihuang,  liujuhua, dubo\}@whu.edu.cn}, \texttt{dacheng.tao@gmail.com}
}

\begin{document}
\maketitle

\input{Section/0_abstract}
\input{Section/1_introduction}
\input{Section/2_related_work}
\input{Section/3_method}

\input{Section/4_experiments}

\input{Section/5_conclusion}
\input{Section/6_limitations}

\section*{Ethics Statement} 
We take ethical considerations very seriously and strictly adhere to the ACL Ethics Policy. 
This paper proposes a method to boost the tool-calling performance of LLMs in long-context scenarios. This method involves both training-free and training-based strategies, using a divide-and-conquer mechanism to assist model decision-making, rather than encouraging them to learn privacy-sensitive knowledge that may raise ethical concerns.
Moreover, all training and evaluation datasets used in this paper are publicly available and have been widely adopted by researchers. Thus, we believe that this research will not pose ethical issues.

\bibliography{acl2026}

\input{Section/7_appendix}

\end{document}

%% file: Section/0_abstract.tex
\begin{abstract}
{\zqh Tool-calling empowers Large Language Models (LLMs) to interact with external environments. However, current methods often struggle to handle massive and noisy candidate tools in long-context tool-calling tasks, limiting their real-world application. To this end, we propose \textbf{Tool-DC}, a \textbf{D}ivide-and-\textbf{C}onquer framework for boosting tool-calling performance of LLMs. The core of Tool-DC is to reduce the reasoning difficulty and make full use of self-reflection ability of LLMs via a ``Try-Check-Retry'' paradigm. Specifically, Tool-DC involves two variants: 1) the training-free Tool-DC (TF), which is plug-and-play and flexible; 2) the training-based Tool-DC (TB), which is more inference-efficient. Extensive experiments show that both Tool-DC methods outperform their counterparts by a clear margin. Tool-DC (TF) brings up to \textbf{+25.10\%} average gains against the baseline on BFCL and ACEBench benchmarks, while Tool-DC (TB) enables Qwen2.5-7B to achieve comparable or even better performance than proprietary LLMs, \textit{e.g.}, OpenAI o3 and Claude-Haiku-4.5.
}

\end{abstract}

%% file: Section/1_introduction.tex
\section{Introduction}
\label{sec:intro}

{\zqh While Large Language Models (LLMs) have shown remarkable successes~\citep{ouyang2022training,achiam2023gpt,touvron2023llama}, they often struggle to deal with the latest information and suffer from factual hallucinations, due to static parameters and closed knowledge boundaries. In response to this issue, the ``\textit{tool-calling}'' paradigm has emerged, \textit{i.e.}, the integration of external tools and application programming interfaces (APIs) enables LLMs to tackle complex, real-world scenarios~\citep{schick2023toolformer,qin2024toolllm}.
}


\begin{figure}[tbp!]
    \centering
    \includegraphics[width=\columnwidth]{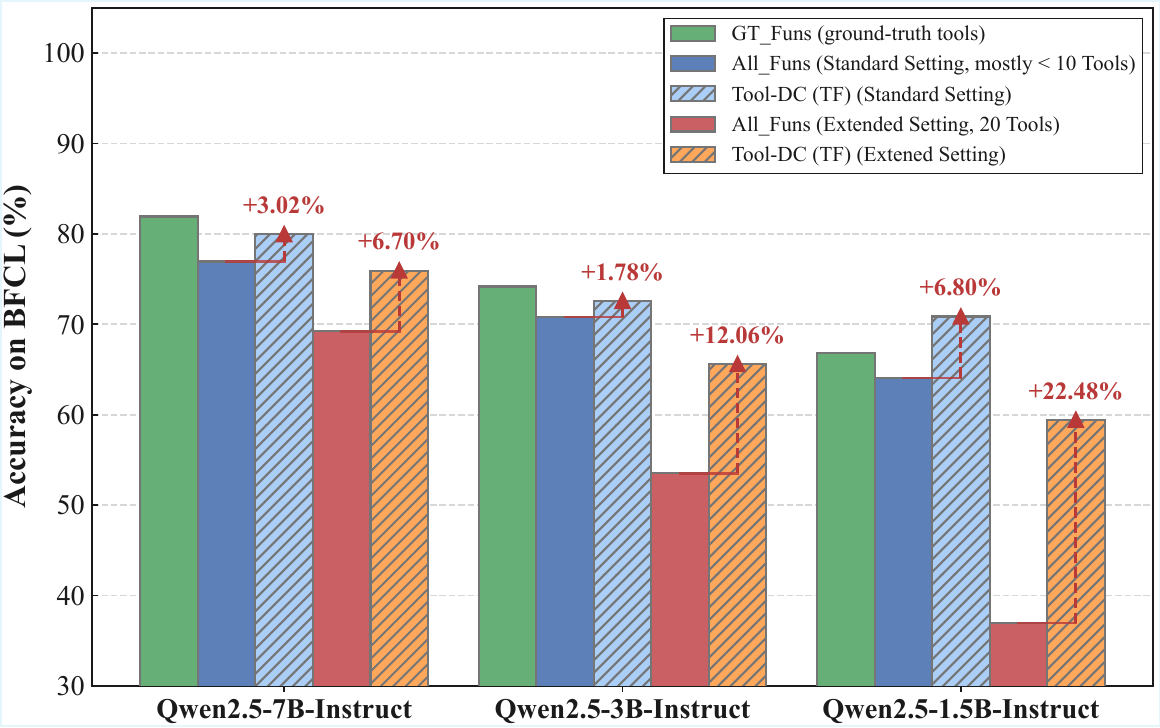}
    \caption{
    {\zqh \textbf{Performance comparison on BFCL~\cite{patilberkeley} with different candidate tool scales.} We see that as the number of candidate tools increases, all models' performance degrades significantly, whereas our Tool-DC method can effectively mitigate this issue.}
    }
    \label{fig:problem-discovery}
\end{figure}

{\zqh The common tool-calling involves several processes. Specifically, given a large number of candidate tools, LLMs need to first select the appropriate tools and then formulate precise input arguments. Intuitively, the performance of tool-calling is sensitive to the number of candidate tools. As shown in our preliminary experiments (Figure~\ref{fig:problem-discovery}), when scaling the number of candidate tools from <10 to 20, all LLMs suffer from performance degradation, especially for the smaller models. We conjecture that, on the one hand, \ding{182} \textit{LLMs may fall short in dealing with long context information} caused by a large number of candidate tools. On the other hand, \ding{183} \textit{confusing tools with similar semantics but different argument descriptions} could affect the argument-filling performance of LLMs.}

{\zqh There are some existing works involving analyzing and addressing these problems. For the first problem, a straightforward approach is to leverage an extra retriever to select the related tool subset~\cite{qin2024toolllm,patel2025dynamic}. While effectively reducing the length of context, it highly relies on the performance of extra retriever~\cite{robertson2009probabilistic,chen2024bge}. More specifically, when the retriever fails in selecting the golden tools, LLMs struggle to predict the correct results due to the absence of golden tools in the context. For the second problem, \citet{cui-etal-2025-enhancing-toolImproving} reveal that many tool-calling errors can be efficiently identified by the structured error checklists. Thus, they manually construct a global error checklist and use it to guide the precise tool-calling via an In-Context Learning (ICL)~\cite{brown2020language} manner. Although effective, the manually defined error checklist is not flexible and is difficult to cover all errors. Thus, there arises a question: \textit{can we explore a more effective and efficient framework to boost the long-context tool-calling performance of LLMs?}
}

{\zqh To achieve this goal, we propose \textbf{Tool-DC}, a \textbf{D}ivide-and-\textbf{C}onquer framework for better LLMs tool-calling, which contains two variants: 1)~a training-free approach, Tool-DC~(TF) and 2)~a training-based approach, Tool-DC~(TB). The core of Tool-DC is a \textit{Try-Check-Retry} paradigm. Specifically, in the processes of Tool-DC~(TF), we first split the total candidate tools into several groups and perform parallel tool-calling (\textit{\textbf{Try}}), then verify the correctness of parallel inference results via strict schema constraints (\textit{\textbf{Check}}), and lastly aggregate the validated results to refine global tool-calling accuracy (\textit{\textbf{Retry}}). By doing so, we can not only reduce the length of context and reasoning difficulty, but also take full advantage of self-reflection capabilities of LLMs for better tool-calling. Moreover, considering the inference latency and deployment costs of Tool-DC~(TF), we further propose to internalize the ``Try-Check-Retry'' decision capabilities into the model parameters. In practice, in Tool-DC~(TB), we first construct a high-quality Chain-of-Thought (CoT) dataset by collecting the correct reasoning traces during the stage of Tool-DC~(TF), and then use it to fine-tune the LLMs.
}

{\zqh We conducted extensive experiments on two widely-used benchmarks, \textit{i.e.}, BFCL~\citep{patilberkeley} and ACEBench~\citep{chen2025acebench}, across multiple LLMs. Notably, considering that the number of candidate tools in the existing benchmarks is generally insufficient (<10), which is inconsistent with real-world scenarios, we introduce an \textit{Extended Setting} by scaling the number of candidate tools to simulate real-world applications. Empirical results show that our Tool-DC~(TF) consistently outperforms the vanilla tool-calling methods among all settings, especially for the small models in the extended setting, \textit{i.e.}, up to \textbf{+25.10\%} average gains. More encouragingly, with the help of Tool-DC~(TB), Qwen2.5-7B~\cite{qwen2025qwen25technicalreport} achieves an \textbf{83.16\%} overall score on BFCL, surpassing the powerful OpenAI o3 and Claude-Haiku-4.5.
}

{\zqh 
\paragraph{Contributions.} To summarize, our contributions are three-fold: (1) We propose a divide-and-conquer framework (Tool-DC) for boosting the tool-calling performance of LLMs in the long-context and confusing tools scenarios.
(2) Tool-DC provides two variants, where the training-free approach is plug-and-play and flexible, and the training-based approach has a higher inference efficiency.
(3) Extensive experiments show that Tool-DC outperforms the vanilla method by a clear margin, \textit{e.g.}, bringing up to \textbf{+25.10\%} average gains against the training-free baseline.
}

%% file: Section/2_related_work.tex
\begin{figure*}[htbp!]
    \centering
    \includegraphics[width=\linewidth]{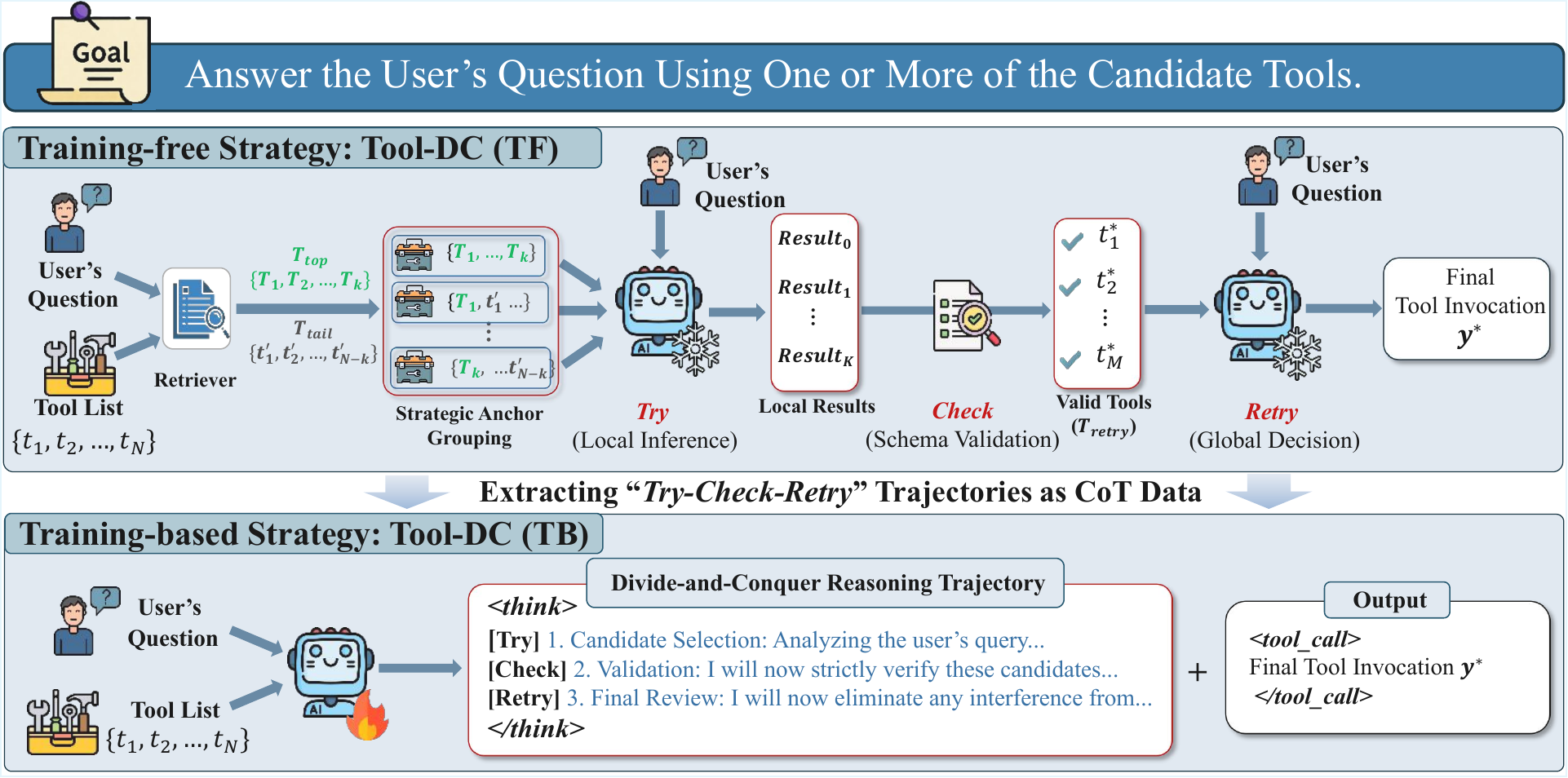}
    \caption{\textbf{Overview of our Tool-DC framework.} The {training-free strategy} employs a ``Try-Check-Retry'' pipeline to reduce the reasoning difficulty, while the {training-based strategy} leverages the prior reasoning trajectories to internalize this divide-and-conquer paradigm into model parameters via fine-tuning.}
\label{fig:framework}
\end{figure*}

\section{Related Works}
\label{sec:related}


Equipping LLMs with external tools extends their capabilities to interact with real-world environments~\citep{qu2025tool,he-etal-2025-gentool,liao-etal-2025-reflectool}. {\zqh However, the manner of providing all tool descriptions in the context is sub-optimal, as some candidate tools could be noisy and redundant~\citep{chen2025enhancing,wang-etal-2025-toolflow}. There are many existing works aiming to improve the performance of tool-calling, which can be classified into training-free and training-based methods.} The former optimize inference via retrieval and constraint injection but remain limited by the capabilities of the frozen base model~\citep{chen2024bge,zheng2024toolrerank,hsieh2023tool,ruan2023tptu,cui-etal-2025-enhancing-toolImproving,dang2025improving}. The latter align models through supervised fine-tuning or reinforcement learning yet often suffer from high training costs and data synthesis bottlenecks~\citep{huang2025toolace,hao2025funreason,zhang2025xlam,qian2025toolrl,zhang2025nemotron}. {\zqh While effectively, these methods still struggle to deal with long-context tool-calling scenarios, where the number of candidate tools is large and many tools are similar but with different argument descriptions. }

{\zqh There are rare works involving addressing the above problems in the tool-calling field~\cite{cui-etal-2025-enhancing-toolImproving,dang2025improving,moon2024efficient}. Among which, \citet{cui-etal-2025-enhancing-toolImproving} attempt to mitigate tool-calling errors via manually designing a global error checklist. Despite its effectiveness, this method needs to carefully design a high-quality checklist, and still struggles with the long-context problem. Different from it, we propose a divide-and-conquer framework, which can effectively reduce the reasoning difficulty and make full use of the self-reflection capabilities of LLMs.
}

%% file: Section/3_method.tex
\section{Methodology}
\label{sec:method}

{\zqh We propose \textbf{Tool-DC}, a Divide-and-Conquer framework for boosting the long-context tool-calling performance of LLMs. As shown in Figure~\ref{fig:framework}, Tool-DC contains two variants: 1) the training-free Tool-DC (TF), which reduces the reasoning difficulty via an explicit ``Try-Check-Retry'' pipeline; 2) the training-based Tool-DC (TB), which internalizes this divide-and-conquer paradigm into model parameters via fine-tuning.}


\subsection{Problem Formulation}
\label{sec:problem}

Let $q$ denote a user query and $\mathcal{T}=\{t_1, t_2, \dots, t_N\}$ represent a library of $N$ candidate tools.
The objective of tool-calling is to enforce the model $\mathcal{M}$ to generate a sequence of tool invocations $\mathbf{y}=\{y_1, y_2, \dots \}$, where each invocation $y_i = (t, \alpha)$ consists of a selected tool $t \in \mathcal{T}$ and its corresponding arguments $\alpha$. A major challenge in real-world scenarios is the scale of $N$. Dealing with a vast search space makes directly modeling $P(\mathbf{y} | q, \mathcal{T})$ ineffective, as the substantial number of irrelevant candidates hinders the model's reasoning capabilities. 
In response to this problem, our Tool-DC aims to partition the global space $\mathcal{T}$ into manageable subspaces $\mathcal{S} = \{S_1, \dots, S_K\}$. This decomposition strategy can reduce the decision complexity and facilitate precise tool-calling.



\subsection{Tool-DC (TF): Training-free Approach}
\label{subsec:training_free}


{\zqh In this part, we introduce the details of our training-free approach, Tool-DC (TF), which involves the following three stages: }

\subsubsection{Try: Grouping and Local Inference}
\label{sec:try_stage}

\begin{figure}[htbp!]
    \centering
    \includegraphics[width=\linewidth]{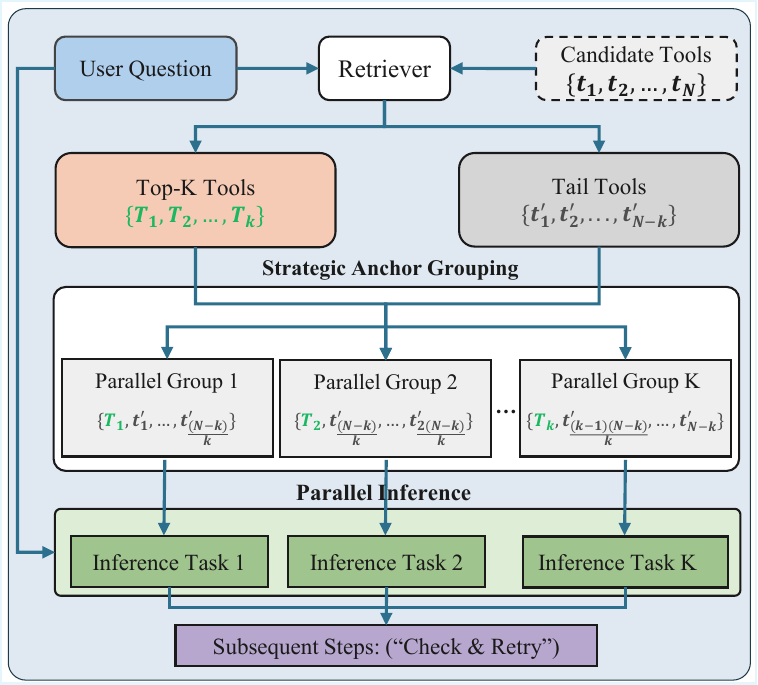}
    \caption{
    {\zqh
    \textbf{Illustration of \textit{Try} stage in Tool-DC (TF)}. By splitting the total candidate tools into several parallel groups, Tool-DC (TF) can reduce the length of context and reasoning difficulty effectively.
    }
    }
\label{fig:try}
\end{figure}

Considering that invoking tools from the global library $\mathcal{T}$ suffers from a low signal-to-noise ratio and interference among tools, in the first stage, we propose to isolate decision-making within low-noise subspaces via Strategic Anchor Grouping and Local Inference, which is illustrated in Figure~\ref{fig:try}.

\paragraph{Strategic Anchor Grouping.} 
{\zqh Similar to many prior works~\cite{qin2024toolllm, patel2025dynamic}, we first retrieve Top-$K$ relevant tools $\mathcal{T}_{top}$ via an extra retriever, \textit{e.g.}, BM25~\cite{robertson2009probabilistic}. Then, we further construct $K$ parallel groups $\{S_1, \dots, S_K\}$ via the Strategic Anchor Grouping strategy. In practice, to obtain the subspace $S_i$, we take the $i$-th tool from $\mathcal{T}_{top}$ as an anchor and add a disjoint subset of distractors from $\mathcal{T}_{tail} = \mathcal{T} \setminus \mathcal{T}_{top}$. Such a grouping method can not only grant tools missed by the initial retrieval (contained in $\mathcal{T}_{tail}$) a chance to be evaluated, but also simultaneously decouple highly relevant candidates to prevent confusion. Notably, we preserve the original Top-$K$ set as a distinct group $S_{0} = \mathcal{T}_{top}$.
}


\paragraph{Local Inference.}
{\zqh For each subspace $S_j \in \{S_0, S_1, \dots, S_K\}$, the model $\mathcal{M}$ performs local inference and outputs an initial tool invocation or a null token:
\begin{equation}
    o_j = \mathcal{M}(q, S_j),
\end{equation}
where $j \in \{0, 1, \dots, K\}$ denotes the index of subspace, and $o_j$ represents the output $(t_j, \alpha_j)$ or the null token $\varnothing$.  This process reduces the search space complexity and enables the model to generate more precise tool arguments.}


\subsubsection{Check: Schema Consistency Validation}
\label{sec:check_stage}

{\zqh Since the generated tool calls usually suffer from hallucinations, \textit{e.g.}, invoking non-existent functions or hallucinating arguments, we leverage a rule-based \textit{Consistency Validator} $\mathcal{F}(\cdot)$ to detect whether the tool calls are valid. Inspired by \citet{cui-etal-2025-enhancing-toolImproving}, we design several schema constraints associated with the tools. In particular, the validity of each predicted output $o_j$ is determined as:
}


\begin{equation}
    \mathbb{I}_{\mathcal{C}}(o_j)=
    \begin{cases}
    1 & \text{if } o_j \neq \varnothing \land \mathcal{F}(o_j, \text{Schema}(S_j)) \\
    0 & \text{otherwise}
    \end{cases},
\end{equation}

{\zqh \noindent 
where $\text{Schema}(\cdot)$ refers to the schema constraints that verify the validity from three dimensions:}
\begin{itemize}
    \item \textbf{Function Name Validity:} The name of invoked function $t_j$ must explicitly exist in the defined tool set.
    \item \textbf{Argument Key Verification:} The keys in $\alpha_j$ must match the defined arguments, and all required parameters must be present.
    \item \textbf{Data Type Consistency:} The values assigned to arguments must adhere to the defined data types (e.g., \texttt{string}, \texttt{integer}, \texttt{boolean}).
\end{itemize}
As a result, this process yields a refined set of valid candidates $\mathcal{V} = \{o_j \mid \mathbb{I}_{\mathcal{C}}(o_j)=1\}$.


\subsubsection{Retry: Global Aggregation and Decision}
\label{sec:retry_stage}

{\zqh 
Lastly, to take advantage of the self-reflection ability of LLMs, we introduce a refinement mechanism that utilizes the above validated candidates to obtain a more precise global decision. Specifically, we collect a refined candidate tool set $\mathcal{T}_{\text{retry}}$ by retrieving the original definitions of the tools presented in the valid set $\mathcal{V}$:
\begin{equation}
    \mathcal{T}_{\text{retry}} = \{ t \in \mathcal{T} \mid \exists (t, \alpha) \in \mathcal{V} \}.
\end{equation}
The refined subset $\mathcal{T}_{\text{retry}}$ is provided in the context, which is then fed into the $\mathcal{M}$ to obtain the final tool invocation $\mathbf{y}^*$:
\begin{equation}
   \mathbf{y}^* = \mathcal{M}(q, \mathcal{T}_{\text{retry}}).
\end{equation}
In general, by splitting the total candidate tools into several subspaces and using the consistency validator to filter the invalid tools, we can obtain a more relevant subset of candidate tools. Furthermore, we employ the prior feedback to enforce the model to self-refine its output, thus leading to better tool-calling performance.
}



\subsection{Tool-DC (TB): Training-based Approach}
\label{subsec:Internalization}

{\zqh Although the proposed Tool-DC (TF) method is plug-and-play and flexible, it requires multiple forward passes and will lead to some inference latency. Thus, as an alternative, we additionally propose a training-based method, Tool-DC (TB), which aims to internalize the divide-and-conquer paradigm into model parameters via fine-tuning.
}


\begin{algorithm}[htbp!]
\small
\caption{Data Construction for Tool-DC (TB)}
\label{alg:data_construction_refined}
\begin{algorithmic}[1]
\State \textbf{Input:} Raw training dataset $\mathcal{D}_{\text{raw}}$, library $\mathcal{T}$, model $\mathcal{M}$
\State \textbf{Output:} CoT training dataset $\mathcal{D}_{\text{CoT}}$
\State $\mathcal{D}_{\text{CoT}} \leftarrow \emptyset$

\For{$(q, \mathbf{y}_{\text{gt}}) \in \mathcal{D}_{\text{raw}}$}
    \State $\mathcal{T}_{\text{valid}} \leftarrow \emptyset$ 
    
    \Statex \textcolor{gray}{\textit{// Phase 1: Try \& Check}}
    \For{$t \in \mathcal{T}$} 
        \State $o \leftarrow \mathcal{M}(q, \{t\})$ \Comment{Local inference}
        \Statex \textcolor{gray}{\quad \quad \textit{// Check: Name, Args, and Type}}
        \If{$\mathbb{I}_{\mathcal{C}}(o) = 1 $}
            \State $\mathcal{T}_{\text{valid}} \leftarrow \mathcal{T}_{\text{valid}} \cup \{t\}$
        \EndIf
    \EndFor
    
    \If{$\mathcal{T}_{\text{valid}} = \emptyset$} \textbf{continue} \EndIf

    \Statex \textcolor{gray}{\textit{// Phase 2: Retry}}
    \State $\mathbf{y^*} \leftarrow \mathcal{M}(q, \mathcal{T}_{\text{valid}})$ 
    
    \Statex \textcolor{gray}{\textit{// Phase 3: Rationale Synthesis}}
    \If{$\mathbf{y^*} = \mathbf{y}_{\text{gt}}$} 
        \State \textbf{Synthesize rationale $R$ via the template}:
  
        \State \parbox[t]{\dimexpr\linewidth-\algorithmicindent}{
        \vspace{-0.5em} 
        \begin{quote}
            \small\itshape\color{gray} 
            ``1. Candidate Selection: ... \\
            2. Validation: ... \\
            3. Final Review: ...''
        \end{quote}
        
    }
        
        \State $\text{CoT} \leftarrow \texttt{<think>} R \texttt{</think>} \texttt{<tool>} y^* \texttt{</tool>}$
        \State $\mathcal{D}_{\text{CoT}} \leftarrow \mathcal{D}_{\text{CoT}} \cup \{(q, \mathcal{T}, \text{CoT})\}$
    \EndIf
\EndFor
\State \Return $\mathcal{D}_{\text{CoT}}$
\end{algorithmic}
\end{algorithm}

{\zqh 
\paragraph{Data Construction.} The key of Tool-DC (TB) is to construct the CoT training data. In practice, we mainly follow the pipeline of Tool-DC (TF) to collect the correct reasoning trajectories by using the xlam-function-calling-60k dataset~\cite{zhang2025xlam} as raw training corpus $\mathcal{D}_{\text{raw}}$.To reduce dependence on external retrievers, we replace the group-based approach with an enumeration strategy, \textit{i.e.}, the number of groups is equal to the number of all candidate tools ($K=N$). The pipeline of CoT data construction is shown in Algorithm~\ref{alg:data_construction_refined}. Notably, for ease of illustration, we simplify the reasoning template in the algorithm. The detailed template is shown in Figure~\ref{fig:prompt_train} of Appendix~\ref{sec:appendix_prompts}.

}


{\zqh
\paragraph{Optimization.} After obtaining the CoT training dataset, denoted as $\mathcal{D}_{\text{CoT}}$, we can fine-tune the model $\mathcal{M}$ to minimize the negative log-likelihood of both reasoning traces and final tool invocation:
\begin{equation}
\mathcal{L} = -\sum_{(q, \mathcal{T}) \in \mathcal{D}_{\text{CoT}}} \log P(\text{CoT} \mid q, \mathcal{T}).
\end{equation}
\noindent By doing so, we can finally obtain the model that can self-reflect and generate the precise tool-calling in a single forward pass.
}



%% file: Section/4_experiments.tex
\section{Experiments}
\label{sec:experiments}
\subsection{Experimental Setup}
\label{sec:setting}

\paragraph{Tasks and Datasets.}

{\zqh 
We conduct the main experiments on two representative benchmarks: Berkeley Function-Calling Leaderboard (BFCL)~\citep{patilberkeley} and ACEBench~\citep{chen2025acebench}. For BFCL, following standard evaluation protocols~\citep{zhang2025nemotron,dang2025improving}, we measure performance on \emph{Non-Live} (synthetic) and \emph{Live} (hand-crafted) subsets, respectively. For ACEBench, we adopt the \textit{Normal (en)} split that covers diverse interaction scenarios as the test set. 

Moreover, to assess robustness on different candidate tool scales, we employ two settings: 1) \textbf{\textit{Standard Setting}} incorporates tool lists of the original benchmark to evaluate basic tool-calling capability; 2) \textbf{\textit{Extended Setting}} expands candidates to 20 functions via randomly injecting irrelevant tools, simulating real-world noise. For evaluation, we report the strict Abstract Syntax Tree (AST) exact-match accuracy for both benchmarks. The details of all benchmarks are provided in Appendix~\ref{sec:appendix_benchmarks}.
}

{\zqh
\paragraph{Models.} We mainly evaluate the effectiveness of Tool-DC on Qwen2.5~\cite{qwen2025qwen25technicalreport} family (1.5B/3B/7B) models. To verify the universality of our methods, we also use the Qwen3-4B~\cite{yang2025qwen3}, Llama-3.1/3.2~\cite{dubey2024llama} series (Llama-3.2-1B/-3B, Llama-3.1-8B), Gemma-3~\cite{team2025gemma} series (1B/4B/12B) instruction models, and two closed-source LLMs, \textit{i.e.}, GPT-4o-mini~\cite{hurst2024gpt} and DeepSeek-V3.2~\cite{liu2024deepseek}. During the implementation of Tool-DC (TF), we set the number of groups to up to 5\footnote{The parameter analysis of $K$ can be found in Section~\ref{sec:analysis}.}, \textit{i.e.}, $K=min(5, N)$. For the extra retriever, we use the representative BM25~\cite{robertson2009probabilistic} method. While for the implementation of Tool-DC (TB), we use the public \texttt{LLaMA-Factory}~\cite{zheng2024llamafactory} toolkit as the training codebase. The details of training and inference hyperparameters are shown in Appendix~\ref{sec:train_hyperparams}. 
}



{\zqh 
\noindent \textbf{Compared Methods.}
We compare our Tool-DC with several cutting-edge counterparts. For training-free methods, we use the following baselines: 1) \textit{GT\_Funs}, only using the ground-truth tools as the context; 2) \textit{All\_Funs}, directly using all candidate tools as the context; 3) \textit{Top-$K$}, using the retrieved Top-$K$ relevant tools as the context; 4) \textit{HiTEC-ICL}~\citep{cui-etal-2025-enhancing-toolImproving}, using the manual designed error checklist to guide the precise tool-calling; 5) \textit{ToolGT (Prompting)}~\citep{dang2025improving}, using a curriculum-inspired prompt to enforce step-by-step tool-calling. For training-based methods, in addition to the base model and vanilla supervised fine-tuning (SFT) baseline, \textit{i.e.}, directly optimizing the model on ground-truth tool invocation without any reasoning traces, we also compare with several proprietary and tool-specialized LLMs. Due to space limitations, we provide the details of all compared models in Appendix~\ref{sec:appendix_details_of_baselines}.

}



\input{Tables/tab1_training_free_main_result}

\begin{figure*}[t]
    \centering
    \includegraphics[width=\linewidth]{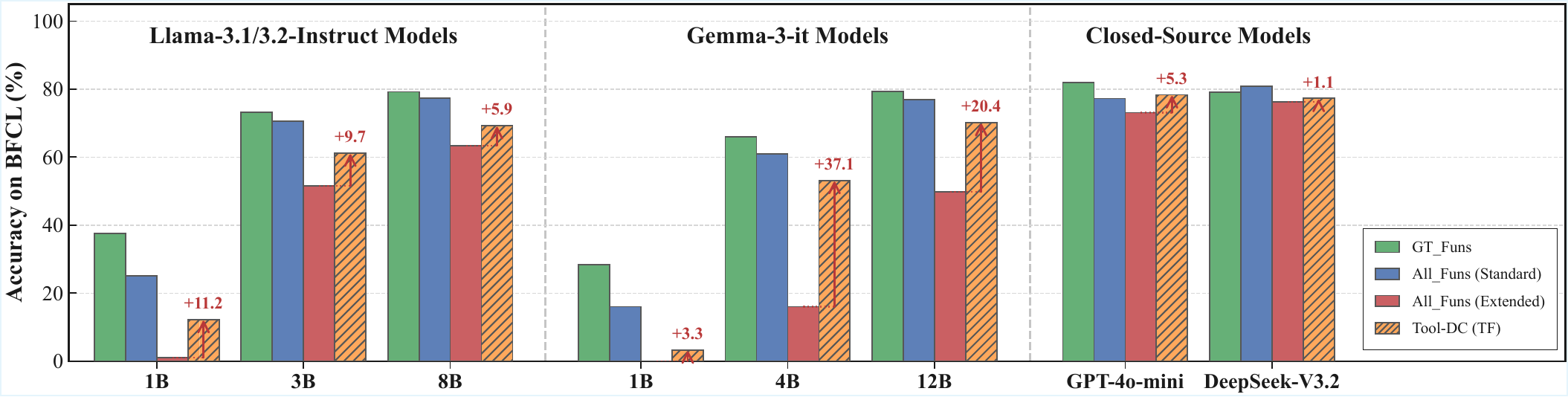}
    \caption{
    {\zqh \textbf{Performance comparison of other LLMs} with different training-free strategies on the BFCL benchmark. Here, we mainly compare our Tool-DC (TF) method with the All\_Funs baseline.}
    }
    \label{fig:figure6}
\end{figure*}

\subsection{Main Results}

\subsubsection{Performance of Training-free Strategy}

{\zqh 
We first evaluate Qwen2.5 family models using our Tool-DC (TF) on BFCL and ACEBench benchmarks, and present the comparative results in Table~\ref{tab:training_free}. From these results, we can find that:
}



{\zqh
\noindent \textbf{Tool-DC (TF) outperforms the baselines across model scales.} First, in the Standard Setting, Tool-DC (TF) achieves the highest average scores across all Qwen2.5 family models, especially for the smaller models. Specifically, on the Qwen2.5-1.5B model, our method achieves an average score of 59.81\%, outperforming the vanilla method by a significant margin, \textit{i.e.}, +4.61\% average gains. This indicates that smaller models struggle more to deal with long-context tool-calling tasks, while our methods can effectively reduce the reasoning difficulty and bring better performance.
}

{\zqh
\noindent \textbf{Tool-DC (TF) exhibits strong robustness in noisy long-context tool-calling scenarios.} 
Compared to the Standard Setting, we introduce more noisy candidate tools in the Extended Setting. As seen, in such a more difficult setting, all methods show performance degradation, especially in smaller models. Notably, for the All\_Funs, the performance degradation is up to 25.72\% average score. Conversely, by locating the relevant tools, our Tool-DC (TF) can alleviate the negative effect of noisy tools and maintain model performance effectively. Specifically, it outperforms the All\_Funs by +25.10\% average score in the Qwen2.5-1.5B model. These results prove the robustness of Tool-DC (TF).
}

{\zqh
\noindent \textbf{Tool-DC (TF) brings consistent performance gains for various LLMs.} In addition to Qwen2.5 models, we also evaluate Tool-DC (TF) on Llama3, Gemma3 and two closed-source LLMs, \textit{i.e.}, GPT-4o-mini and DeepSeek-V3.2. Figure~\ref{fig:figure6} illustrates the comparative results on the BFCL benchmark, from which we find that our Tool-DC (TF) consistently boosts the tool-calling performance across all base models. Notably, for the powerful GPT-4o-mini model, Tool-DC (TF) can still lead to +5.3\% performance gains, proving its effectiveness.
}


\input{Tables/tab2_traning_base_main_result_bfcl}

\begin{figure}[t]
    \centering
    
    \includegraphics[width=\columnwidth]{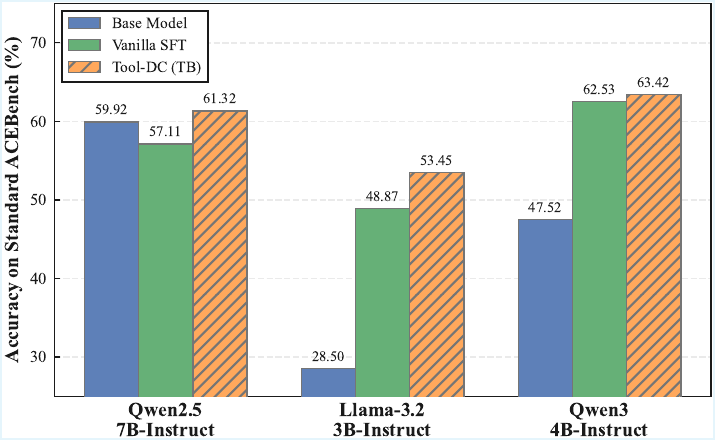}
    \caption{{\zqh
    Comparison between base models and tuned models using Tool-DC (TB) on Standard ACEBench.
    }
    }
    \label{fig:acebench}
\end{figure}

\subsubsection{Performance of Training-based Strategy}
{\zqh 
The comparative results of different tuned models on ACEBench and BFCL benchmarks are presented in Figure~\ref{fig:acebench} and Table~\ref{tab:train_base1}, respectively. Notably, due to limited computing resources, we only adopt Tool-DC (TB) to fine-tune some representative models. From them, we can observe that:
}


{\zqh 
\noindent \textbf{\textbf{Tool-DC (TB) yields consistent gains across diverse models.}} In addition to the results on Qwen2.5 family models, we also apply our Tool-DC (TB) to boost the tool-calling performance of Llama3 and Qwen3 models. Taking the ACEBench benchmark as an example, we find that Tool-DC (TB) brings +24.95\% performance gain for Llama-3.2-3B-Instruct model, and +15.9\% for Qwen3-4B-Instruct model. These results can prove the universality of our Tool-DC (TB) method.
}



{\zqh
\noindent \textbf{\textbf{Our tuned models achieve remarkable performance against the proprietary LLMs.}} 
In Table~\ref{tab:train_base1}, we report the official results from the leaderboard\footnote{https://gorilla.cs.berkeley.edu/leaderboard.html} of BFCL for the proprietary LLMs. It can be seen, our Tool-DC (TB) can not only improve the tool-calling performance for various base models, but also enable them to achieve remarkable (or even better) performance against the other proprietary LLMs. More specifically, with the help of Tool-DC (TB), Qwen2.5-7B-Instruct achieves 83.16\% overall score on BFCL, outperforming the OpenAI o3, DeepSeek-V3.2, and Claude-Haiku-4.5. These results demonstrate the superiority of our method.
}


\input{Tables/ablation}

\subsection{Ablation Study}
\label{sec:ablation}

{\zqh
In this part, we conduct ablation studies to investigate the effect of different stages (\textit{Try}, \textit{Check}, and \textit{Retry}) of Tool-DC (TF) method, respectively. Notably, we use the Qwen2.5-3B-Instruct as the base model, and report the results on the Extended Setting of BFCL in Table~\ref{tab:ablation}. As seen, removing any stage in Tool-DC (TF) will lead to performance degradation, proving their effectiveness. 
}

{\zqh
Firstly, when removing the \textit{Try} stage, the average performance of Qwen2.5-3B-Instruct drops from 64.77\% to 36.79\%, indicating the importance of task decomposition. Without the divide-and-conquer mechanism, Tool-DC (TF) falls short in checking its outputs and self-reflection. Secondly, when removing the \textit{Check} stage, the refinement process tends to be unstable and lead to sub-optimal results. Lastly, when removing the \textit{Retry} stage, we observe a performance collapse, \textit{i.e.}, from 64.77\% to 5.26\%. This highlights: after filtering out potentially correct answers via \textit{Check}, Tool-DC (TF) relies on \textit{Retry} to leverage these positive signals. Without \textit{Retry}, Tool-DC (TF) cannot process the high-potential candidates identified by \textit{Check}, resulting in the loss of valid solutions.
}

\begin{figure}[t]
    \centering
    \includegraphics[width=\columnwidth]{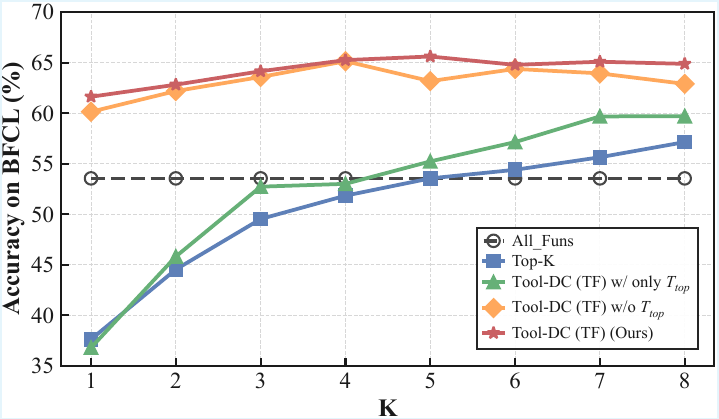}
    \caption{\textbf{Impact of Group Count $K$}. We evaluate Qwen2.5-3B-Instruct using different training-free methods on the Extended Setting of BFCL. 
    }
    \label{fig:impact_k}
\end{figure}

\subsection{More Analyses}
\label{sec:analysis}

{\zqh
\paragraph{Sensitivity analysis of Group Count $K$.}
We investigate the impact of the group count $K$, which is an important hyperparameter in the first stage of Tool-DC (TF). In practice, we conduct comparative experiments on the Extended Setting of BFCL by ranging $K$ from 1 to 8. As shown in Figure~\ref{fig:impact_k}, the Top-$K$ baseline method suffers from significant performance drops at smaller $K$ (<3) due to aggressive sample filtering. Conversely, by making full use of all useful candidate tools, our Tool-DC (TF) exhibits superior robustness to the number of $K$. More specifically, in Figure~\ref{fig:impact_k}, ``Tool-DC (TF) w/ only $\mathcal{T}_{top}$'' denotes that we only use the retrieved Top-$K$ tools in the \textit{Try} stage, which can be seen as an improvement on the Top-$K$ method via our \textit{Check} and \textit{Retry} stages. By doing so, Tool-DC (TF) consistently improves the performance of the Top-$K$ baseline, indicating the effectiveness of the self-reflection mechanism. Moreover, when $K$ is set to 5, our Tool-DC (TF) method performs best, thus leaving it as the default setting in our work.
}

\begin{figure}[t]
    \centering
    \includegraphics[width=0.985\columnwidth]{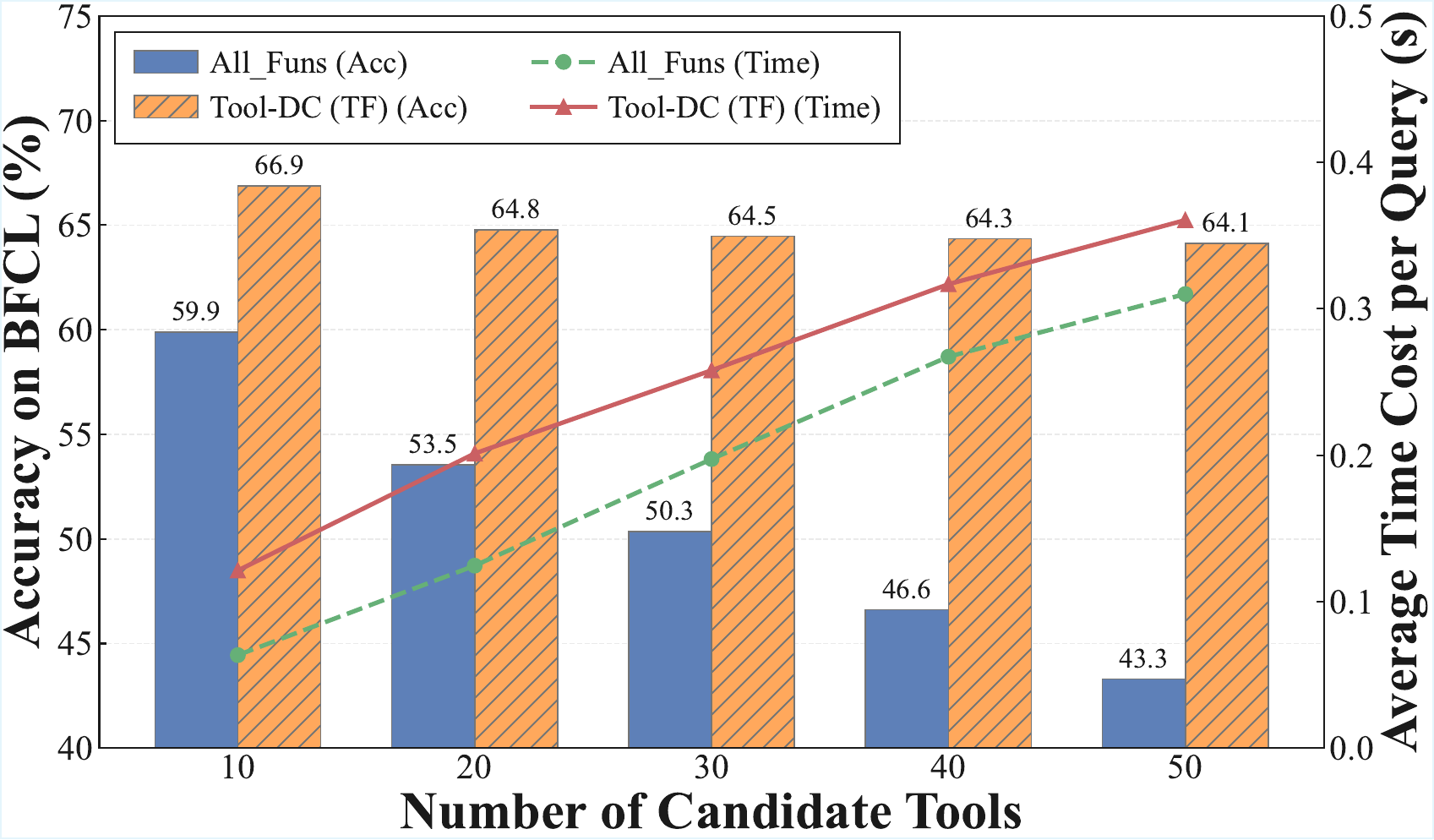}
    \caption{
    \textbf{Scalability and Efficiency Analysis.} Here, we use the Qwen2.5-3B-Instruct as the base model, and compare our Tool-DC (TF) with the All\_Funs baseline method on the Extended Setting of BFCL.
    }
    \label{fig:efficiency}
\end{figure}


{\zqh
\paragraph{Scalability and efficiency analysis.}
To better simulate the real-world scenarios where the candidate tools are massive and noisy, we increase the number of candidate tools $N$ from 10 to 50 for the BFCL benchmark by randomly adding some irrelevant tools. We compare the results of the Qwen2.5-3B-Instruct model by using the All\_Funs and our Tool-DC (TF), respectively. As illustrated in Figure~\ref{fig:efficiency}, the performance of All\_Funs baseline drops significantly as the $N$ increases, \textit{i.e.}, from 59.88\% to 43.30\%. On the contrary, our Tool-DC (TF) can maintain the performance effectively, demonstrating its robustness. Furthermore, we additionally compare the inference latency between Tool-DC (TF) and All\_Funs. Although Tool-DC (TF) indeed leads to some inference overhead, it is tolerable against the performance gains of Tool-DC (TF).



}


%% file: Tables/tab1_training_free_main_result.tex
\begin{table*}[t]
\centering
\small
\setlength{\tabcolsep}{3.5pt}
\caption{
    \textbf{Performance comparison of Qwen2.5 models} with different training-free strategies on BFCL and ACEBench. The best and second-best results are in \textbf{bold} and \underline{underlined}, respectively. The subscript results denote the relative performance gains against the All\_Funs baseline.
    }
\label{tab:training_free}
\resizebox{\textwidth}{!}{%
\begin{tabular}{llcccclccccl}
\toprule
 &  & \multicolumn{5}{c}{\textbf{Standard Setting}} & \multicolumn{5}{c}{\textbf{Extended Setting}} \\
\cmidrule(lr){3-7} \cmidrule(lr){8-12}
\multicolumn{2}{l}{\textbf{Methods}} & \multicolumn{3}{c}{BFCL} & ACEBench & \multirow{2}{*}{\textbf{Avg.}} & \multicolumn{3}{c}{BFCL} & ACEBench & \multirow{2}{*}{\textbf{Avg.}} \\
\cmidrule(lr){3-5} \cmidrule(lr){8-10}
 &  & Non-Live & Live & Overall & Overall &  & Non-Live & Live & Overall & Overall &  \\
\midrule
\rowcolor{gray!20} \multicolumn{12}{l}{\textit{\textbf{Qwen2.5-1.5B-Instruct}}} \\
 & GT\_Funs & 70.96 & 62.77 & 66.87 & 47.92 & 57.40 & 70.96 & 62.77 & 66.87 & 47.92 & 57.40 \\
 & All\_Funs & 69.96 & 58.18 & 64.07 & 46.33 & 55.20 & 40.69 & 33.23 & 36.96 & 22.00 & 29.48 \\
 & Top-$K$ & 76.35 & 63.80 & 70.08 & 44.52 & 57.30\gain{2.10} & 60.31 & 43.60 & 51.96 & 38.58 & 45.27\gain{15.79} \\
 & HiTEC-ICL & \underline{75.56} & \underline{61.51} & \underline{68.54} & 45.17 & \underline{56.86}\gain{1.66} & \underline{46.46} & \underline{39.97} & \underline{43.22} & 25.42 & 34.32\gain{4.84} \\
 & ToolGT (Prompting) & 65.29 & 52.26 & 58.78 & \underline{47.50} & 53.14\loss{2.06} & 39.58 & 35.83 & 37.71 & \underline{35.33} & \underline{36.52}\gain{7.04} \\
\rowcolor[RGB]{233,246,255}  & \textbf{Tool-DC (TF) (Ours)} & \textbf{77.63} & \textbf{63.43} & \textbf{70.53} & \textbf{49.08} & \textbf{59.81}\gain{4.61} & \textbf{64.13} & \textbf{62.00} & \textbf{63.07} & \textbf{46.08} & \textbf{54.58}\gain{25.10} \\
\midrule
\rowcolor{gray!20} \multicolumn{12}{l}{\textit{\textbf{Qwen2.5-3B-Instruct}}} \\
 & GT\_Funs & 79.23 & 69.13 & 74.18 & 54.42 & 64.30 & 79.23 & 69.13 & 74.18 & 54.42 & 64.30 \\
 & All\_Funs & 78.08 & 63.51 & 70.80 & 47.25 & 59.03 & 62.73 & 44.34 & 53.54 & 36.50 & 45.02 \\
 & Top-$K$ & 78.00 & 66.03 & 72.02 & 45.78 & 58.90\loss{0.13} & 63.73 & 46.71 & 55.22 & 38.02 & 46.62\gain{1.60} \\
 & HiTEC-ICL & \textbf{78.90} & \underline{63.80} & \underline{71.35} & 49.33 & \underline{60.34}\gain{1.31} & \underline{63.79} & 45.23 & 54.51 & 34.92 & 44.72\loss{0.30} \\
 & ToolGT (Prompting) & 68.94 & 57.22 & 63.08 & \textbf{50.50} & 56.79\loss{2.24} & 59.29 & \underline{54.92} & \underline{57.11} & \underline{46.58} & \underline{51.85}\gain{6.83} \\
 \rowcolor[RGB]{233,246,255} & \textbf{Tool-DC (TF) (Ours)} & \underline{78.54} & \textbf{66.62} & \textbf{72.58} & \underline{49.42} & \textbf{61.00}\gain{1.97} & \textbf{71.79} & \textbf{57.74} & \textbf{64.77} & \textbf{48.17} & \textbf{56.47}\gain{11.45} \\ 
\midrule
\rowcolor{gray!20} \multicolumn{12}{l}{\textit{\textbf{Qwen2.5-7B-Instruct}}} \\
 & GT\_Funs & 85.81 & 78.02 & 81.92 & 65.17 & 73.55 & 85.81 & 78.02 & 81.92 & 65.17 & 73.55 \\
 & All\_Funs & \textbf{86.46} & 67.44 & 76.95 & 59.92 & 68.44 & 75.71 & 62.77 & 69.24 & 58.58 & 63.91 \\
 & Top-$K$ & 84.60 & 75.50 & 80.05 & 50.83 & 65.44\loss{3.00} & 70.23 & 54.18 & 62.21 & 46.31 & 54.26\loss{9.65} \\
 & HiTEC-ICL & \underline{84.98} & \textbf{76.24} & \textbf{80.61} & 61.33 & \underline{70.97}\gain{2.53} & \underline{76.81} & \underline{65.14} & \underline{70.98} & 54.67 & 62.83\loss{1.08} \\
 & ToolGT (Prompting) & 77.73 & 68.02 & 72.88 & \textbf{68.92} & 70.90\gain{2.46} & 76.00 & 60.55 & 68.28 & \textbf{62.42} & \underline{65.35}\gain{1.44} \\
 \rowcolor[RGB]{233,246,255} & \textbf{Tool-DC (TF) (Ours)} & 84.73 & \underline{74.83} & \underline{79.78} & \underline{64.00} & \textbf{71.89}\gain{3.45} & \textbf{84.40} & \textbf{70.00} & \textbf{77.20} & \underline{58.83} & \textbf{68.02}\gain{4.11} \\
\bottomrule
\end{tabular}
}
\end{table*}

%% file: Tables/tab2_traning_base_main_result_bfcl.tex

\begin{table}[t]
    \centering
    \caption{ 
    {\zqh \textbf{Performance comparison between our tuned models and other proprietary models on BFCL}. Notably, the results of proprietary models are from the official leaderboard of BFCL.}
    }
    \label{tab:train_base1}
    
    \renewcommand{\arraystretch}{1.2}
    \setlength{\tabcolsep}{4pt} 
    
    \footnotesize 
    
    \begin{tabularx}{\columnwidth}{X c c c}
        \toprule
        \textbf{Models} & \textbf{\makecell[b]{Non-Live}} & \textbf{\makecell[b]{Live}} & \textbf{\makecell[b]{Overall}} \\
        
        \midrule
        
        \rowcolor{gray!20} \multicolumn{4}{l}{\textbf{Close-source Models}} \\
        OpenAI o3 & 81.94 & 73.21 & 77.58  \\
        DeepSeek-V3.2-Exp & 85.52 & 76.02 & 80.77 \\
        Claude-Haiku-4.5 & 86.50 & 78.68 	& 82.59  \\
        Gemini-3-Pro-Preview & \textbf{90.65} & \textbf{83.12} & \textbf{86.89} \\
        \midrule
        \rowcolor{gray!20} \multicolumn{4}{l}{\textbf{Open-source Models}} \\
        \multicolumn{4}{l}{\textit{\textbf{Large-scale Models}}} \\
        
        \hspace{0.5em}Qwen3-235B-A22B-Inst & 90.12 & 76.61 & 83.37 \\
        \hspace{0.5em}Llama-3.3-70B-Inst & 88.02 & 76.61 & 82.32 \\
        \hdashline
        \multicolumn{4}{l}{\textit{\textbf{Tool-specialized Models}}} \\
        \hspace{0.5em}xLAM-2-8b-fc-r & 84.58 & 67.95 & 76.27 \\
        \hspace{0.5em}ToolACE-DEV-8B & 89.67 & 75.20 & 82.44 \\
        \hspace{0.5em}ToolACE-MT-8B & 84.94 & 71.52 & 78.23 \\
        \hspace{0.5em}Hammer2.1-3B & 85.50 & 69.50 & 77.50 \\
        \hspace{0.5em}Hammer2.1-7B & 84.96 & 70.54 & 77.75 \\
        \midrule
        
        \rowcolor{gray!20} \multicolumn{4}{l}{\textbf{Tuned Models}} \\
        Qwen2.5-7B-Instruct & 86.46 & 67.44 & 76.95 \\
        \quad -w Vanilla SFT & 84.23 & 74.24 & 79.24\gain{2.29} \\
        \rowcolor[RGB]{233,246,255} \quad -w Tool-DC (TB) & 86.67 & 79.64 & 83.16\gain{6.21} \\ \hdashline
        Llama-3.2-3B-Instruct & 82.67 & 58.33 & 70.50 \\
        \quad -w Vanilla SFT & 83.31 & 62.99 & 73.15\gain{2.65} \\
        \rowcolor[RGB]{233,246,255} \quad -w Tool-DC (TB) & 83.17 & 63.95 & 73.56\gain{3.06} \\
        

        \bottomrule
    \end{tabularx} 
    
\end{table}

%% file: Tables/ablation.tex
\begin{table}[t]
    \caption{\label{tab:ablation} \textbf{Ablation study on Tool-DC (TF)}. Here, we use the Qwen2.5-3B-Instruct as the base model and evaluate on the Extended Setting of BFCL.
    }
    \centering
    \resizebox{\columnwidth}{!}{
        \begin{tabular}{l ccc ccc}
        \toprule
        \multirow{2}{*}{\textbf{Methods}} & \multicolumn{3}{c}{\textbf{Modules}} & \multicolumn{3}{c}{\textbf{Performance}} \\
        \cmidrule(lr){2-4} \cmidrule(lr){5-7}
         &\small Try &\small Check &\small Retry &\small Non-Live &\small Live &\small Overall \\
        \midrule
        
        Base & \xmark & \xmark & \xmark & 62.73 & 44.34 & 53.54 \\
        
        \textbf{Tool-DC (TF)} & \cmark & \cmark & \cmark & \textbf{71.79} & \textbf{57.74} & \textbf{64.77} \\
 
        \midrule
        
        \textit{w/o} Try & \xmark & \cmark & \cmark & 44.19 & 29.39 & 36.79 \\
        
        \textit{w/o} Check & \cmark & \xmark & \cmark & 52.06 & 52.41 & 52.24 \\
        
        \textit{w/o} Retry & \cmark & \cmark & \xmark & 7.33 & 3.18 & 5.26 \\
        
        \bottomrule
        \end{tabular}
    }
\end{table}

%% file: Section/5_conclusion.tex
\section{Conclusion}
\label{sec:conclusion}


{\zqh
In this paper, we propose Tool-DC, a divide-and-conquer framework to boost the tool-calling performance of LLMs via mitigating the negative effect of long-context and noisy candidate tools. Specifically, Tool-DC involves two variants: the training-free Tool-DC (TF) and training-based Tool-DC (TB). The former approach leverages the ``Try-Check-Retry'' pipeline to reduce the reasoning difficulty and make full use of the self-reflection ability of LLMs, thus leading to precise tool-calling. The latter approach aims to internalize the divide-and-conquer paradigm into model parameters via fin-tuning. Extensive experiments show that our Tool-DC consistently outperforms both training-free and training-based counterparts across various model scales and architectures.

}

%% file: Section/6_limitations.tex
\section*{Limitations}
\label{sec:limitations}

{\zqh
Our work has several potential limitations that present opportunities for future research.
First, we only used xlam-function-calling-60k as the seed dataset to construct our CoT data for Tool-DC (TB). Although this dataset is large, it lacks sufficient diversity and contains a limited number of tools in the context. In the future, we plan to construct a more diverse training set with massive and noisy context to simulate real-world scenarios, and explore reinforcement learning, \textit{e.g.}, GRPO~\cite{guo2025deepseek}, to further improve the model performance.
Secondly, our current experiments are limited to single-step tool-calling tasks, and we do not evaluate our framework on multi-step nested scenarios, such as those introduced in the latest version of BFCL. Therefore, future work could extend our method to more complex tool-calling settings.
}


%% file: Section/7_appendix.tex
\appendix
\section{Appendix}
\label{sec:appendix}

\subsection{Details of Benchmarks}
\label{sec:appendix_benchmarks}

\paragraph{\textbf{The Berkeley Function-Calling Leaderboard (BFCL)}.} \citet{patilberkeley} introduce a comprehensive benchmark BFCL for evaluating function-calling capabilities of LLMs, comprising two primary evaluation categories: Non-Live and Live.
Each category further includes four distinct task types: single, multiple, parallel, and parallel multiple function invocations. The total number of samples in the used test set is 2,501.
 
\begin{itemize} 
\item \textbf{Single}: LLMs are provided with a single function description and must generate exactly one corresponding call. 
\item \textbf{Multiple}: LLMs are provided with a set of candidate functions and must select and invoke the most appropriate one. 
\item \textbf{Parallel}: LLMs need to generate multiple function calls simultaneously for a single query, determining the appropriate number of invocations. 
\item \textbf{Parallel Multiple}: Combining the challenges of \textit{Parallel} and \textit{Multiple}, LLMs need to select from several candidates to invoke zero or more functions as required.
\end{itemize}

\paragraph{\textbf{ACEBench}.} 
ACEBench~\cite{chen2025acebench} is a diverse tool-calling benchmark that offers fine-grained coverage across a broad range of domains.
Specifically, it encompasses 8 major domains and 68 sub-domains, including but not limited to technology, finance, entertainment, society, health, culture, and environment.
The benchmark is organized into three primary categories: {Normal}, {Special}, and {Agent}.
In this work, we focus exclusively on the \textbf{Normal} category, which contains 828 test samples.
The Normal category is further subdivided into the following five tasks: \textit{Single-Turn}, \textit{Multi-Turn}, \textit{Similar APIs}, \textit{Profile}, and \textit{Atom}.

\subsection{Training and Inference Details}
\label{sec:train_hyperparams}
In this work, we use the xlam-function-calling-60k dataset~\cite{zhang2025xlam} as raw training corpus, and construct a CoT training set with 46,897 samples. Table~\ref{tab:hyperparams} shows the detailed training hyperparameters. Specifically, we set the batch size to 16 and the peak learning rate to 2e-5. All models are trained with LoRA~\cite{hulora} for 2 epochs on two NVIDIA A800 80GB GPUs, requiring approximately 10 hours for finetuning 7B models using our Tool-DC (TB). For inference, we use the VLLM\footnote{https://github.com/vllm-project/vllm} engine for accelerating the model generation, and set the temperature to 0, the maximum length of output tokens to 512.  

\begin{table}[h]
    \centering
    
    \caption{Training hyperparameters for all models.}
    \label{tab:hyperparams}
    \resizebox{0.6\columnwidth}{!}{
        \begin{tabular}{lc}
            \toprule
            \textbf{Hyperparameter} & \textbf{Value} \\
            \midrule
            Batch Size  & 16 \\
            Learning Rate & 2e-5 \\
            LR Scheduler & cosine \\
            Warmup Ratio & 0.1 \\
            Epochs & 2 \\
            \bottomrule
        \end{tabular}
    }
\end{table}


\subsection{Details of Compared Models}
\label{sec:appendix_details_of_baselines}
In our training-based experiments, we compare our tuned models with both general-purpose 
commercial APIs (OpenAI o3\footnote{https://openai.com/index/introducing-o3-and-o4-mini}~\cite{Openai-o3}, DeepSeek-V3.2-Exp\footnote{https://api-docs.deepseek.com/news/news250929}~\cite{liu2025deepseek}, Claude-Haiku-4.5\footnote{https://www.anthropic.com/news/claude-haiku-4-5}~\cite{claude-haiku-4-5}, and Gemini-3-Pro-Preview\footnote{https://deepmind.google/models/gemini/pro/}~\cite{gemini3})
and open-source large-scale models (Qwen3-235B-A22B-Inst~\cite{yang2025qwen3} and Llama-3.3-70B-Inst~\cite{dubey2024llama}), as well as some specialized tool-calling models, including ToolACE-DEV-8B~\cite{huang2025toolace}, ToolACE-MT-8B~\cite{zeng2025toolace}, xLAM-2~\cite{zhang2025xlam}, and Hammer2.1(1.5B/3B/7B)~\cite{lin2024hammer}.

\subsection{Sensitivity Analysis of Group Count $K$}
\label{sec:appendix_k}
\begin{figure}[t]
    \centering
    \includegraphics[width=0.85\linewidth]{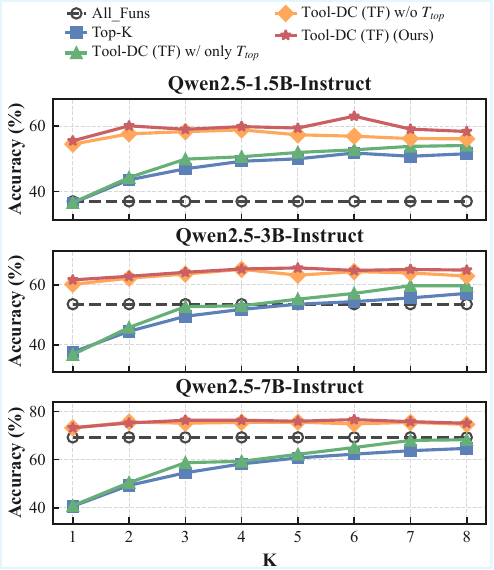}
    \caption{
    Sensitivity analysis of group count $K$ across Qwen2.5 series. All models are evaluated on the Extended Setting of BFCL.
    }
    \label{fig:impact_k_multi}
\end{figure}

We extend the sensitivity analysis of group count $K$ to Qwen2.5-1.5B, 3B, and 7B-Instruct. Results in Figure~\ref{fig:impact_k_multi} consistently show that while baselines like \textit{Top-K} suffer significant drops at low $K$ ($<3$) due to over-filtering, our method remains robust even at $K=1$. Performance stabilizes at $K \in [4, 6]$, confirming that a moderate partition size effectively balances noise reduction and context completeness across varying model scales.

\input{Tables/tab_longcontext_model_result}

\subsection{Performance on Long-context LLMs}

To verify whether the long-context tool-calling issue can be fixed by scaling the context windows, we evaluate the cutting-edge long-context LLMs, including Qwen2.5-7B-Instruct-1M~\cite{yang2025qwen2} and InternLM2.5-7B-Chat-1M~\cite{cai2024internlm2}, which have up to 1M tokens in the context windows, on the BFCL benchmark. Specifically, we mainly conduct experiments on the Extended Settings, and report the comparative results in Table~\ref{tab:long_context_model_result}. For reference, we also report the results of GT\_Funs as the upper-bound. 

As seen, while InternLM2.5-7B-Chat-1M has a long context window, it still suffers from dramatic performance drops in the Extended Setting, indicating that simply scaling context windows cannot address the long-context tool-calling problem. This underscores the necessity of exploring a more robust tool-calling method. Encouragingly, with the help of our Tool-DC (TF), InternLM2.5-7B-Chat-1M can alleviate the performance drop effectively, \textit{i.e.}, from 16.29\% to 38.70\%, continuing to prove its effectiveness and superiority.


\input{Tables/more_retriever}
\subsection{Performance on More Retrievers}
\label{sec:appendix_more_retrievers}
In our work, we use the representative unsupervised retrieval method BM25 as the base retriever. Here, we further conduct experiments to investigate the generality of Tool-DC on the other retrievers. Specifically, two cutting-edge dense retrievers, including bge-reranker-v2-m3~\cite{chen2024bge} and ToolBench\_IR~\cite{qin2024toolllm}, are used in this experiment. We evaluate the Qwen2.5-3B-Instruct model on the Extended Setting of BFCL and report the comparative results in Table~\ref{tab:more_retriever_result}. From these results, we can find that our Tool-DC (TF) brings consistent performance gains against the Top-$K$ baseline across all base retrievers, proving its generality. Notably, although bge-reranker-v2-m3 is effective, it is a two-tower BERT~\cite{devlin2019bert} model, which is inference-inefficient. While ToolBench\_IR is finetuned on ToolBench, it is difficult to generalize to other tool-calling tasks, thus leading to sub-optimal results. Moreover, considering BM25 is a powerful and efficient unsupervised retriever~\cite{lei2023unsupervised}, we finally choose it as the base retriever in our experiment.


\input{Tables/prompt}
\subsection{Prompt Details}
\label{sec:appendix_prompts}

Here, we provide the detailed prompts used in our Tool-DC framework. Specifically, for Tool-DC (TF), the ``Try'' and ``Retry'' prompts are presented in Figure~\ref{fig:prompt_try} and Figure~\ref{fig:prompt_retry}, respectively. For Tool-DC (TB), we present the prompt template of data construction in Figure~\ref{fig:prompt_train}.


\subsection{Full Results}
\label{sec:appendix_detail_result}
Due to space limitations, we only report the averaged experimental results in the main body of this paper. Here, we present the detailed results to help readers better reproduce our main experiments. Specifically, the detailed results of training-free approaches on BFCL and ACEBench benchmarks are provided in Table~\ref{tab:detail_train_free_bfcl} and Table~\ref{tab:detail_train_free_ace}, respectively. The detailed results of training-based approaches on BFCL and ACEBench benchmarks are provided in Table~\ref{tab:bfcl_sft} and Table~\ref{tab:ace_sft}, respectively.

\subsection{AI Assistant Usage}
During the paper writing, we used a proprietary LLM, \textit{i.e.}, Gemini-3-Pro, as the general-purpose assistant to polish some sentences. We state that we did not prompt them to generate research ideas or search for related works, thus avoiding the risk of introducing false information.


\input{Tables/tab_detail_train_free_bfcl}
\input{Tables/tab_detail_train_base_bfcl}

\input{Tables/tab_detail_train_free_ace}
\input{Tables/tab_detail_train_base_ace}

%% file: Tables/tab_longcontext_model_result.tex
\begin{table*}[t]
\centering
\small
\setlength{\tabcolsep}{3pt} 
\renewcommand{\arraystretch}{1.1} 

\caption{Performance of long-context LLMs on the Extended Setting of BFCL.
}
\label{tab:long_context_model_result}

\resizebox{\textwidth}{!}{%
\begin{tabular}{lccccccccccc}
\toprule
\multirow{2}{*}{\textbf{Methods}} & \multicolumn{4}{c}{\textbf{Non-Live}} & \multicolumn{4}{c}{\textbf{Live}} & \multicolumn{3}{c}{\textbf{Overall}} \\
\cmidrule(lr){2-5} \cmidrule(lr){6-9} \cmidrule(lr){10-12}
 & Simple & Multiple & Parallel & \begin{tabular}{@{}c@{}}Parallel\\Multiple\end{tabular} & Simple & Multiple & Parallel & \begin{tabular}{@{}c@{}}Parallel\\Multiple\end{tabular} & Non-Live & Live & Overall \\
\midrule

\rowcolor{gray!20} \multicolumn{12}{l}{\textbf{Qwen2.5-7B-Instruct-1M}} \\
\quad GT\_Funs   & 74.92 & 95.00 & 90.50 & 85.50 & 80.23 & 77.30 & 62.50 & 66.67 & 86.48 & 77.50 & 81.99 \\
\quad All\_Funs & 68.67 & 85.50 & 84.00 & 71.00 & 63.18 & 65.81 & 62.50 & 58.33 & 77.29 & 65.14 & 71.22 \\
\rowcolor[RGB]{233,246,255} \quad \textbf{Tool-DC (TF) (Ours)}  & 73.17  &89.00 	&88.50 	&79.50  &65.89  &69.71 &62.50 	&66.67 	&82.54 	&68.84 	&75.69 \\
\midrule

\rowcolor{gray!20} \multicolumn{12}{l}{\textbf{InternLM2.5-7B-Chat-1M}} \\
\quad GT\_Funs   & 51.92 & 85.00 & 70.50 & 51.00 & 55.81 & 51.28 & 25.00 & 20.83 & 64.60 & 51.30 & 57.95 \\
\quad All\_Funs&17.92 &26.00 & 25.00 	&10.50 	&14.34 	&12.44 	&18.75 	&4.17 	&19.85 	&12.73 	&16.29 \\
\rowcolor[RGB]{233,246,255} \quad \textbf{Tool-DC (TF) (Ours)}  & 41.00 & 65.00 & 46.00 & 28.50  & 35.66  & 32.19 & 12.50 & 12.50 & 45.12 	& 32.27 & 38.70 \\

\bottomrule
\end{tabular}%
}
\end{table*}

%% file: Tables/more_retriever.tex
\begin{table*}[htbp!]
\centering
\small
\setlength{\tabcolsep}{3pt} 
\renewcommand{\arraystretch}{1.1} 

\caption{Performance of Qwen2.5-3B-Instruct using different retrievers on the Extended Setting of BFCL.
}
\label{tab:more_retriever_result}

\resizebox{\textwidth}{!}{%
\begin{tabular}{lccccccccccc}
\toprule
\multirow{2}{*}{\textbf{Methods}} & \multicolumn{4}{c}{\textbf{Non-Live}} & \multicolumn{4}{c}{\textbf{Live}} & \multicolumn{3}{c}{\textbf{Overall}} \\
\cmidrule(lr){2-5} \cmidrule(lr){6-9} \cmidrule(lr){10-12}
 & Simple & Multiple & Parallel & \begin{tabular}{@{}c@{}}Parallel\\Multiple\end{tabular} & Simple & Multiple & Parallel & \begin{tabular}{@{}c@{}}Parallel\\Multiple\end{tabular} & Non-Live & Live & Overall \\
\midrule

 GT\_Funs & 70.92 & 91.50 & 79.00 & 75.50 & 67.83 	& 70.37 & 56.25 & 37.50 & 79.23 & 69.13 &74.18 \\

All\_Funs &64.42  &71.00  &62.50  &53.00  &43.41 	&44.92 	&43.75 	&29.17 	&62.73 	&44.34 	&53.54 
 \\
\midrule
\rowcolor{gray!20} \multicolumn{12}{l}{\textbf{bge-reranker-v2-m3
}} \\
\quad -w  Top-$K$ &67.42 &83.50 &74.50 	&61.00 	&52.71 	&64.10 	&50.00 	&37.50 	&71.60 	&61.29 	&66.45 \\

\rowcolor[RGB]{233,246,255} \quad -w Tool-DC (TF) &68.08 &82.50 	&75.50 	&68.50 	&57.75 	&64.29 	&50.00 	&41.67 	&73.65 	&62.47 	&68.06  \\
\midrule
\rowcolor{gray!20} \multicolumn{12}{l}{\textbf{{ToolBench\_IR}}} \\
\quad -w Top-$K$ & 63.42 & 73.00 & 66.00 & 41.50 & 42.25 & 47.10 & 50.00 & 37.50 & 60.98 & 46.04 & 53.51 \\
\rowcolor[RGB]{233,246,255} \quad -w Tool-DC (TF) & 68.83 & 81.50 & 74.00 & 58.50 & 53.10 & 62.01 & 56.25 & 37.50 & 70.71 & 59.81 & 65.26 \\
\midrule
\rowcolor{gray!20} \multicolumn{12}{l}{\textbf{BM25}} \\
\quad -w Top-$K$ & 64.42 & 74.50 & 71.00 & 45.00 & 41.09 & 48.53 & 43.75 & 29.17 & 63.73 & 46.71 & 55.22 \\
\rowcolor[RGB]{233,246,255} \quad -w Tool-DC (TF) & 64.67 & 81.00 & 77.50 & 64.00 & 48.84 & 60.40 & 43.75 & 45.83 & 71.79 & 57.74 & 64.77 \\

\bottomrule
\end{tabular}%
}
\end{table*}

%% file: Tables/prompt.tex
\definecolor{aclblue}{RGB}{46, 117, 182}
\definecolor{acldark}{RGB}{30, 30, 30}
\definecolor{aclbg}{RGB}{247, 250, 255}
\definecolor{codegreen}{RGB}{68, 163, 64}
\definecolor{codepurple}{RGB}{128, 0, 128}
\definecolor{codeorange}{RGB}{230, 100, 0}

\newtcolorbox{promptbox}[2][]{
    enhanced,
    title={\textbf{#2}},
    colframe=aclblue,
    colback=aclbg,
    colbacktitle=aclblue,
    coltitle=white,
    fonttitle=\bfseries\small,
    fontupper=\small\ttfamily,
    boxrule=0.8pt,
    arc=2mm,
    left=3mm, right=3mm, top=3mm, bottom=3mm,
    width=\linewidth,
    #1
}

\newcommand{\role}[1]{\textcolor{aclblue}{\textbf{#1}}}
\newcommand{\placeholder}[1]{\textcolor{codegreen}{<#1>}}

\begin{figure*}[t] %
    \centering
    \begin{promptbox}{Prompt template of ``Try'' stage in Tool-DC (TF)}
        \role{\# System:}\\
        You are a Function Selection Expert. Your task is to identify ALL functions that are semantically relevant to the user's question from the provided list. Extract information from the user's question and substitute it into the function parameters.
        
        Read the user's question and the function descriptions carefully. Choose any function that could potentially meet user needs or meet a part of user needs.
        
        If you decide to invoke any of the function(s), you MUST put it in the format of: 
        \textcolor{codeorange}{[func\_name1(params\_name1=params\_value1...), func\_name2(params)]}. You SHOULD NOT include any other text in the response.\\
        Here is a list of functions in json format that you can invoke.
        
        \placeholder{Tools of Each Group}

        \role{\# User:}\\
        \placeholder{Question}
    \end{promptbox}
    \caption{The prompt template of ``Try'' stage in Tool-DC (TF).}
    \label{fig:prompt_try}
\end{figure*}

\begin{figure*}[t]
    \centering
    \begin{promptbox}{Prompt template of ``Retry'' stage in Tool-DC (TF)}
        \role{\# System:}\\
        You are an expert in composing functions. You are given a question and a set of possible functions. Based on the question, you will need to make one or more function/tool calls to achieve the purpose. If none of the functions can be used, point it out. If the given question lacks the parameters required by the function, also point it out.
        
        You should only return the function calls in your response.
        If you decide to invoke any of the function(s), you MUST put it in the format of:
        \textcolor{codeorange}{[func\_name1(params\_name1=params\_value1...), func\_name2(params)]}. 
        You SHOULD NOT include any other text in the response.
        
        At each turn, you should try your best to complete the tasks requested by the user within the current turn. Continue to output functions to call until you have fulfilled the user's request to the best of your ability.
        
        Here is a list of functions in json format that you can invoke.
        \placeholder{Tools}
        
        \role{\# User:}\\
        \placeholder{Question}
    \end{promptbox}
    \caption{The prompt template of ``Retry'' stage in Tool-DC (TF).}
    \label{fig:prompt_retry}
\end{figure*}

\begin{figure*}[t]
    \centering
    \begin{promptbox}{Prompt template for constructing the CoT data}
        \role{\# System:}\\
        You are an expert in composing functions. You are given a question and a set of possible functions. Based on the question, you will need to make one or more function/tool calls to achieve the purpose. If none of the function can be used, point it out. If the given question lacks the parameters required by the function, also point it out. 
        
        You should only return the function call in tools call sections. Continue to output functions to call until you have fulfilled the user's request to the best of your ability.
 
        Here is a list of functions in json format that you can invoke.
        \placeholder{Tools}
                      
        For each interaction, you MUST strictly follow this two-step process:

        \textbf{Step 1: Reasoning (\textless think\textgreater)} \\
        You must engage in a detailed chain-of-thought enclosed within \textcolor{codeorange}{\textless think\textgreater \textless/think\textgreater} tags. 
        
        This process must follow these exact 3 sub-steps: \\
        \textbf{Candidate Selection:} Analyze the user's question. Iterate through the function list... \\
        \textbf{Validation:} Strictly check the candidate list against function definitions... \\
        \textbf{Final Review:} Focus exclusively on the effective candidate list... \\
        \textbf{Step 2: Invoke (\textless tool\_call\textgreater)} \\
        If you decide to invoke function(s), output them in the following specific format:
        \textcolor{codeorange}{\textless tool\_call\textgreater [func\_name1(params\_name1=value1, ...), func\_name2(params)]\textless/tool\_call\textgreater}
        
        \role{\# User:}\\
        \placeholder{Question}

        \role{\# Assistent:}\\
        \textcolor{codeorange}{\textless think\textgreater}
        
    \textbf{1. Candidate Selection:} Analyzing the user's query, I will attempt to map key information to the function parameters. The functions that potentially match and may have their parameters filled are: \placeholder{candidate\_tool\_calls} 
    
    \textbf{2. Validation:} I will now strictly verify these candidates against their definitions, ensuring all parameter types and constraints are met. The functions that pass this strict verification are: \placeholder{valid\_tools}. 
    
    \textbf{3. Final Review:} I will now eliminate any interference from irrelevant functions and focus solely on the valid candidates. 
    
    \textcolor{codeorange}{\textless /think\textgreater}
    
    \textcolor{codeorange}{\textless tool\_call\textgreater}\placeholder{final\_tool\_invocation}\textcolor{codeorange}{\textless /tool\_call\textgreater}    
    \end{promptbox}
    \caption{The prompt template for constructing the CoT data in Tool-DC (TB).
    }
    \label{fig:prompt_train}
\end{figure*}

%% file: Tables/tab_detail_train_free_bfcl.tex
\begin{table*}[htbp!]
\centering
\small
\setlength{\tabcolsep}{2.5pt}
\caption{
    {Detailed results of different training-free methods on BFCL}, which is a full version of Table~\ref{tab:training_free}.
    }
\label{tab:detail_train_free_bfcl}
\resizebox{\textwidth}{!}{%
\begin{tabular}{llcccclcccclccc}
\toprule
 & & \multicolumn{4}{c}{\textbf{Non-Live}} & & \multicolumn{4}{c}{\textbf{Live}} & & \multicolumn{3}{c}{\textbf{Overall}} \\
\cmidrule(lr){3-6} \cmidrule(lr){8-11} \cmidrule(lr){13-15}
\multicolumn{2}{l}{\textbf{Methods}} & Simple & Multiple & Parallel & \begin{tabular}{@{}c@{}}Parallel\\Multiple\end{tabular} & & Simple & Multiple & Parallel & \begin{tabular}{@{}c@{}}Parallel\\Multiple\end{tabular} & & Non-Live & Live & Overall \\
\midrule
\multicolumn{15}{c}{\textbf{Standard Setting}} \\
\midrule
\rowcolor{gray!20} \multicolumn{15}{l}{\textit{\textbf{Qwen2.5-1.5B-Instruct}}} \\

& GT\_Funs & 63.33 & 88.50 & 68.50 & 63.50 & & 65.12 & 63.25 & 43.75 & 29.17 & & 70.96 & 62.77 & 66.87 \\

 & All\_Funs & 63.33 & 84.00 & 68.00 & 64.50 & & 65.12 & 57.26 & 43.75 & 33.33 & & 69.96 & 58.18 & 64.07 \\
 & Top-$K$ & 66.42 & \underline{83.50} & \underline{79.50} & \underline{76.00} & & \underline{68.99} & \underline{63.15} & \underline{50.00} & \underline{45.83} & & \underline{76.35} & \textbf{63.80} & \underline{70.08} \\
 & HiTEC-ICL & \textbf{67.75} & \underline{83.50} & 79.00 & 72.00 & & \textbf{70.54} & 59.83 & \textbf{56.25} & 41.67 & & 75.56 & 61.51 & 68.54 \\
 & ToolGT (Prompting) & \underline{66.67} & 78.00 & 64.00 & 52.50 & & 67.05 & 49.29 & 25.00 & 41.67 & & 65.29 & 52.26 & 58.78 \\
 \rowcolor[RGB]{233,246,255} & \textbf{Tool-DC (TF) (Ours) } & 66.50 & \textbf{85.50} & \textbf{80.50} & \textbf{78.00} & & 68.60 & 62.68 & \underline{50.00} & \textbf{50.00} & & \textbf{77.63} & \underline{63.43} & \textbf{70.53} \\
\midrule
\rowcolor{gray!20} \multicolumn{15}{l}{\textit{\textbf{Qwen2.5-3B-Instruct}}} \\

& GT\_Funs & 70.92 & 91.50 & 79.00 & 75.50 & & 67.83 & 70.37 & 56.25 & 37.50 & & 79.23 & 69.13 & 74.18 \\

 & All\_Funs & 70.83 & 88.00 & 79.50 & 74.00 & & 68.22 & 63.06 & 56.25 & 37.50 & & 78.08 & 63.51 & 70.80 \\
 & Top-$K$ & \underline{71.00} & \underline{89.50} & 78.00 & \underline{73.50} & & \underline{67.05} & \underline{66.29} & \textbf{62.50} & \underline{45.83} & & 78.00 & \underline{66.03} & \underline{72.02} \\
 & HiTEC-ICL & \textbf{72.58} & 87.50 & \textbf{80.50} & \textbf{75.00} & & \textbf{68.60} & 63.25 & \textbf{62.50} & 37.50 & & \textbf{78.90} & 63.80 & 71.35 \\
 & ToolGT (Prompting) & 67.75 & 81.50 & 68.50 & 58.00 & & \textbf{68.60} & 55.46 & 25.00 & 33.33 & & 68.94 & 57.22 & 63.08 \\
 \rowcolor[RGB]{233,246,255} & \textbf{Tool-DC (TF) (Ours) } & 70.67 & \textbf{90.00} & \underline{78.50} & \textbf{75.00} & & \underline{67.05} & \textbf{66.95} & \textbf{62.50} & \textbf{50.00} & & \underline{78.54} & \textbf{66.62} & \textbf{72.58} \\
\midrule
\rowcolor{gray!20} \multicolumn{15}{l}{\textit{\textbf{Qwen2.5-7B-Instruct}}} \\

& GT\_Funs & 72.25 & 96.00 & 91.00 & 84.00 & & 77.13 & 78.73 & 62.50 & 66.67 & & 85.81 & 78.02 & 81.92 \\

 & All\_Funs & 75.33 & 94.50 & 91.50 & 84.50 & & 76.74 & 74.93 & 62.50 & 70.83 & & 86.46 & 67.44 & 76.95 \\
 & Top-$K$ & \underline{71.92} & 93.50 & \textbf{91.00} & \underline{82.00} & & 77.13 & \underline{75.40} & \textbf{62.50} & \underline{70.83} & & 84.60 & \underline{75.50} & \underline{80.05} \\
 & HiTEC-ICL & \textbf{72.42} & \textbf{95.00} & 89.00 & \textbf{83.50} & & \textbf{78.29} & \textbf{76.07} & \textbf{62.50} & \underline{70.83} & & \textbf{84.98} & \textbf{76.24} & \textbf{80.61} \\
 & ToolGT (Prompting) & 69.92 & 82.50 & 84.00 & 74.50 & & \textbf{78.29} & 65.53 & 56.25 & \textbf{75.00} & & 77.73 & 68.02 & 72.88 \\
 \rowcolor[RGB]{233,246,255} & \textbf{Tool-DC (TF) (Ours) } & \underline{71.92} & \underline{94.00} & \textbf{91.00} & \underline{82.00} & & 77.91 & 74.93 & \textbf{62.50} & 66.67 & & \underline{84.73} & 75.20 & 79.97 \\
\midrule
\multicolumn{15}{c}{\textbf{Extended Setting}} \\
\midrule
\rowcolor{gray!20} \multicolumn{15}{l}{\textit{\textbf{Qwen2.5-1.5B-Instruct}}} \\

& GT\_Funs & 63.33 & 88.50 & 68.50 & 63.50 & & 65.12 & 63.25 & 43.75 & 29.17 & & 70.96 & 62.77 & 66.87 \\

 & All\_Funs & 51.75 & 49.00 & 35.50 & 26.50 & & 32.95 & 33.90 & 25.00 & 12.50 & & 40.69 & 33.23 & 36.96 \\
 & Top-$K$ & \underline{61.75} & \textbf{69.50} & \underline{69.50} & \textbf{40.50} & & 40.70 & \underline{45.20} & 25.00 & \underline{16.67} & & \underline{60.31} & \underline{43.60} & \underline{51.96} \\
 & HiTEC-ICL & 57.33 & 53.00 & 45.50 & 30.00 & & \underline{41.09} & 40.36 & \underline{31.25} & \underline{16.67} & & 46.46 & 39.97 & 43.22 \\
 & ToolGT (Prompting) & 46.83 & 55.00 & 31.00 & 25.50 & & 38.76 & 36.37 & 0.00 & 4.17 & & 39.58 & 35.83 & 37.71 \\
 \rowcolor[RGB]{233,246,255} & \textbf{Tool-DC (TF) (Ours) } & \textbf{64.13} & \underline{59.50} & \textbf{76.50} & \underline{40.00} & & \textbf{76.50} & \textbf{67.00} & \textbf{53.50} & \textbf{64.13} & & \textbf{64.13} & \textbf{62.00} & \textbf{63.07} \\
\midrule
\rowcolor{gray!20} \multicolumn{15}{l}{\textit{\textbf{Qwen2.5-3B-Instruct}}} \\

& GT\_Funs & 70.92 & 91.50 & 79.00 & 75.50 & & 67.83 & 70.37 & 56.25 & 37.50 & & 79.23 & 69.13 & 74.18 \\

 & All\_Funs & 64.42 & 71.00 & 62.50 & 53.00 & & 43.41 & 44.92 & 43.75 & 29.17 & & 62.73 & 44.34 & 53.54 \\
 & Top-$K$ & \underline{64.42} & \underline{74.50} & \underline{71.00} & 45.00 & & 41.09 & 48.53 & \textbf{43.75} & 29.17 & & 63.73 & 46.71 & 55.22 \\
 & HiTEC-ICL & 64.17 & 71.50 & 64.00 & \underline{55.50} & & 44.19 & 45.68 & \textbf{43.75} & \underline{37.50} & & \underline{63.79} & 45.23 & 54.51 \\
 & ToolGT (Prompting) & 58.67 & 71.50 & 54.50 & 52.50 & & \textbf{56.59} & \underline{55.37} & \textbf{43.75} & 25.00 & & 59.29 & \underline{54.92} & \underline{57.11} \\
 \rowcolor[RGB]{233,246,255} & \textbf{Tool-DC (TF) (Ours) } & \textbf{64.67} & \textbf{81.00} & \textbf{77.50} & \textbf{64.00} & & \underline{48.84} & \textbf{60.40} & \textbf{43.75} & \textbf{45.83} & & \textbf{71.79} & \textbf{57.74} & \textbf{64.77} \\
\midrule
\rowcolor{gray!20} \multicolumn{15}{l}{\textit{\textbf{Qwen2.5-7B-Instruct}}} \\

& GT\_Funs & 72.25 & 96.00 & 91.00 & 84.00 & & 77.13 & 78.73 & 62.50 & 66.67 & & 85.81 & 78.02 & 81.92 \\

 & All\_Funs & 65.33 & 82.00 & 85.00 & 70.50 & & 58.14 & 63.91 & 75.00 & 54.17 & & 75.71 & 62.77 & 69.24 \\
 & Top-$K$ & 67.42 & \underline{81.00} & 83.50 & 49.00 & & 48.45 & 55.94 & 50.00 & 41.67 & & 70.23 & 54.18 & 62.21 \\
 & HiTEC-ICL & 65.75 & \textbf{86.00} & \underline{85.00} & \underline{70.50} & & \underline{62.79} & \underline{65.72} & \underline{68.75} & 62.50 & & \underline{76.81} & \underline{65.14} & \underline{70.98} \\
 & ToolGT (Prompting) & \underline{70.00} & 80.00 & 82.50 & \textbf{71.50} & & 58.91 & 61.06 & 43.75 & \underline{66.67} & & 76.00 & 60.55 & 68.28 \\
 \rowcolor[RGB]{233,246,255} & \textbf{Tool-DC (TF) (Ours) } & \textbf{84.40} & 71.58 & \textbf{94.75} & 50.00 & & \textbf{91.00} & \textbf{91.00} & \textbf{84.00} & \textbf{84.40} & & \textbf{84.40} & \textbf{70.00} & \textbf{77.20} \\
\bottomrule
\end{tabular}
}
\end{table*}

%% file: Tables/tab_detail_train_base_bfcl.tex
\begin{table*}[htbp!]
\centering
\small
\setlength{\tabcolsep}{3pt} 
\renewcommand{\arraystretch}{1.1} 

\caption{
    {Detailed results of different training-based methods on BFCL}, which is a full version of Table~\ref{tab:train_base1}.
}
\label{tab:bfcl_sft}

\resizebox{\textwidth}{!}{%
\begin{tabular}{lccccccccccc}
\toprule
\multirow{2}{*}{\textbf{Methods}} & \multicolumn{4}{c}{\textbf{Non-Live}} & \multicolumn{4}{c}{\textbf{Live}} & \multicolumn{3}{c}{\textbf{Overall}} \\
\cmidrule(lr){2-5} \cmidrule(lr){6-9} \cmidrule(lr){10-12}
 & Simple & Multiple & Parallel & \begin{tabular}{@{}c@{}}Parallel\\Multiple\end{tabular} & Simple & Multiple & Parallel & \begin{tabular}{@{}c@{}}Parallel\\Multiple\end{tabular} & Non-Live & Live & Overall \\
\midrule

\textbf{Qwen2.5-7B-Instruct} & 75.33 & 94.50 & 91.50 & 84.50 & 76.74 & 74.93 & 62.50 & 70.83 & 86.46 & 67.44 & 76.95 \\
\quad -w Vanilla SFT  & 70.92 & 95.50 & 88.50 & 82.00 & 77.52 & 73.69 & 81.25 & 58.33 & 84.23 & 74.24 & 79.24 \\

\rowcolor[RGB]{233,246,255} \quad -w Tool-DC (TB)   & 71.17 & 95.00 & 93.00 & 87.50 & 83.72 & 79.58 & 56.25 & 54.17 & 86.67 & 79.64 & \textbf{83.16} \\

\midrule






\textbf{Llama-3.2-3B-Instruct} & 70.67 & 92.50 & 88.50 & 79.00 & 58.33 & 65.12 & 57.64 & 25.00 & 82.67 & 58.33 & 70.50 \\
\quad -w Vanilla SFT  & 71.25 & 92.00 & 88.50 & 81.50 & 70.93 & 61.35 & 56.25 & 54.17 & 83.31 & 62.99 & 73.15 \\

\rowcolor[RGB]{233,246,255} \quad -w Tool-DC (TB)  & 71.67 & 89.50 & 89.50 & 82.00 & 68.60 & 64.01 & 18.75 & 41.67 & 83.17 & 63.95 & \textbf{73.56} \\


\bottomrule
\end{tabular}%
}
\end{table*}

%% file: Tables/tab_detail_train_free_ace.tex
\begin{table*}[t!]
\centering
\setlength{\tabcolsep}{2.5pt}
\caption{
{Detailed results of different training-free methods on ACEBench}, which is a full version of Table~\ref{tab:training_free}. ``Atom'', ``Single Turn'', ``Multiple Turn'', ``Similar API'', and ``Profile'' are subsets of ACEBench.
}
\label{tab:detail_train_free_ace}
\resizebox{\textwidth}{!}{%
\begin{tabular}{@{}l cccccc cccccc@{}}
\toprule
\multirow{2.5}{*}{\textbf{Methods}} & \multicolumn{6}{c}{\textbf{Standard Setting}} & \multicolumn{6}{c}{\textbf{Extended Setting}} \\

\cmidrule(lr){2-7} \cmidrule(lr){8-13}
 & {Atom} & \makecell{{Single} \\ {Turn}} & \makecell{{Multiple} \\ {Turn}} & \makecell{{Similar} \\ {API}} & {Profile} & {Overall} & {Atom} & \makecell{{Single} \\ {Turn}} & \makecell{{Multiple} \\ {Turn}} & \makecell{{Similar} \\ {API}} & {Profile} & {Overall} \\
\midrule

\rowcolor{gray!20} \multicolumn{13}{l}{\textit{\textbf{Qwen2.5-1.5B-Instruct}}} \\
GT\_Funs & 56.00 & 38.50 & 28.00 & 52.00 & 54.00 & 47.92 & 56.00 & 38.50 & 28.00 & 52.00 & 54.00 & 47.92 \\
All\_Funs & 56.67 & 35.00 & 29.00 & 50.00 & 38.00 & 46.33 & 27.67 & 15.00 & 5.00 & 32.00 & 26.00 & 22.00 \\
Top-$K$ & \underline{57.33} & \underline{37.00} & 10.12 & \underline{52.00} & \underline{44.00} & 44.52 & \underline{52.67} & \underline{28.00} & 8.50 & \textbf{40.00} & 34.00 & \underline{38.58} \\
HiTEC-ICL & 54.33 & 34.00 & \underline{24.00} & \underline{52.00} & \textbf{48.00} & 45.17 & 28.00 & 21.50 & 13.00 & 24.00 & \textbf{44.00} & 25.42 \\
ToolGT (Prompting) & \textbf{60.67} & 33.00 & \textbf{30.00} & 48.00 & 32.00 & \underline{47.50} & 49.00 & 18.00 & \underline{17.00} & \underline{26.00} & 34.00 & 35.33 \\
\rowcolor[RGB]{233,246,255} \textbf{Tool-DC (TF)} & 56.33 & \textbf{47.50} & 23.00 & \textbf{66.00} & \underline{44.00} & \textbf{49.08} & \textbf{57.67} & \textbf{38.50} & \textbf{25.00} & \textbf{40.00} & \underline{40.00} & \textbf{46.08} \\

\midrule

\rowcolor{gray!20} \multicolumn{13}{l}{\textit{\textbf{Qwen2.5-3B-Instruct}}} \\
GT\_Funs & 63.33 & 47.50 & 28.00 & 60.00 & 62.00 & 54.42 & 63.33 & 47.50 & 28.00 & 60.00 & 62.00 & 54.42 \\
All\_Funs & 57.33 & 45.50 & 17.00 & 50.00 & 48.00 & 47.25 & 43.00 & 32.00 & 17.00 & 48.00 & 34.00 & 36.50 \\
Top-$K$ & 53.67 & \underline{49.50} & 11.21 & \textbf{62.00} & 44.00 & 45.78 & 47.33 & \underline{38.50} & 6.61 & \underline{54.00} & 28.00 & 38.02 \\
HiTEC-ICL & 53.33 & \textbf{52.00} & \textbf{30.00} & \underline{58.00} & \underline{50.00} & 49.33 & 39.67 & 33.50 & 19.00 & 50.00 & 26.00 & 34.92 \\
ToolGT (Prompting) & \underline{60.67} & 48.00 & \textbf{30.00} & 56.00 & \textbf{54.00} & \textbf{52.50} & \textbf{56.67} & 36.50 & \underline{22.00} & 52.00 & \textbf{50.00} & \underline{46.58} \\
\rowcolor[RGB]{233,246,255} \textbf{Tool-DC (TF)} & \textbf{61.00} & 40.50 & 23.00 & 50.00 & \underline{50.00} & \underline{49.42} & \underline{55.67} & \textbf{46.00} & \textbf{28.00} & \textbf{60.00} & \underline{36.00} & \textbf{48.17} \\

\midrule

\rowcolor{gray!20} \multicolumn{13}{l}{\textit{\textbf{Qwen2.5-7B-Instruct}}} \\
GT\_Funs & 71.00 & 61.00 & 52.00 & 68.00 & 62.00 & 65.17 & 71.00 & 61.00 & 52.00 & 68.00 & 62.00 & 65.17 \\
All\_Funs & 62.33 & 63.50 & 45.00 & 64.00 & 64.00 & 59.92 & 65.33 & 54.50 & 42.00 & 62.00 & 56.00 & 58.58 \\
Top-$K$ & 61.33 & 58.00 & 0.00 & 64.00 & \textbf{62.00} & 50.83 & 53.67 & 50.50 & 17.37 & 52.00 & 46.00 & 46.31 \\
HiTEC-ICL & 65.33 & 61.00 & 48.00 & \underline{66.00} & 60.00 & 61.33 & 58.67 & 48.00 & 44.00 & 62.00 & \textbf{58.00} & 54.67 \\
ToolGT (Prompting) & \textbf{79.67} & \underline{61.50} & \underline{49.00} & \underline{66.00} & \textbf{62.00} & \textbf{68.92} & \textbf{71.33} & \underline{59.50} & \textbf{53.00} & \underline{64.00} & \underline{56.00} & \textbf{64.42} \\
\rowcolor[RGB]{233,246,255} \textbf{Tool-DC (TF)} & \underline{68.00} & \textbf{66.00} & \textbf{50.00} & \textbf{70.00} & 58.00 & \underline{64.00} & \underline{60.00} & \textbf{65.00} & \underline{48.00} & \textbf{66.00} & 54.00 & \underline{58.83} \\

\bottomrule
\end{tabular}
}
\end{table*}

%% file: Tables/tab_detail_train_base_ace.tex
\begin{table*}[t]
\centering
\footnotesize 
\setlength{\tabcolsep}{4pt} %
\renewcommand{\arraystretch}{1.1} %

\caption{
    {Detailed results of different training-based methods on Standard ACEBench}, referred to Figure~\ref{fig:acebench}.
}
\label{tab:ace_sft}

\resizebox{0.85\linewidth}{!}{%
\begin{tabular}{@{}lcccccc@{}}
\toprule
\textbf{Methods} & \textbf{Atom} & \textbf{\makecell{Single Turn}} & \textbf{\makecell{Multiple Turn}} & \textbf{\makecell{Similar API}} & \textbf{Profile} & \textbf{Overall} \\

\midrule



\textbf{Qwen2.5-7B-Instruct}   & 62.33 & 63.50 & 45.00 & 64.00 & 64.00 & 59.92 \\
\quad -w Vanilla SFT   & 67.67 & 65.50 & 7.14 & 70.00 & 64.00 & 57.11 \\
\rowcolor[RGB]{233,246,255} \quad -w Tool-DC (TB)    & 76.33 & 47.00 & 22.94 & 74.00 & 64.00 & \textbf{61.32} \\
\midrule

\textbf{Llama-3.2-3B-Instruct} & 33.33 & 21.00 & 8.00 & 34.00 & 50.00 & 28.50 \\
\quad -w Vanilla SFT   & 65.00 & 43.00 & 5.22 & 52.00 & 48.00 & 48.87 \\
\rowcolor[RGB]{233,246,255} \quad -w Tool-DC (TB)    & 72.67 & 43.00 & 12.71 & 56.00 & 38.00 & \textbf{53.45} \\
\midrule

\textbf{Qwen3-4B-Instruct-2507} & 55.67 & 43.00 & 22.13 & 52.00 & 54.00 & 47.52 \\
\quad -w Vanilla SFT    & 75.00 & 65.50 & 16.67 & 70.00 & 66.00 & 62.53 \\
\rowcolor[RGB]{233,246,255} \quad -w Tool-DC (TB)     & 67.67 & 59.50 & 52.00 & 68.00 & 64.00 & \textbf{63.42} \\

\bottomrule
\end{tabular}%
}
\end{table*}